\documentclass{article}

\usepackage[preprint]{corl_2024} 

\newcommand{\scientist}{experimenter\xspace}
\newcommand{\Scientist}{Experimenter\xspace}
\title{Game On: \\ Towards Language Models as RL {\Scientist}s}

\author{
  Jingwei Zhang$^*$\\
  Google DeepMind\\
  United Kingdom\\
  \texttt{zhangjingwei@google.com} \\
  \And
  Thomas Lampe$^*$\\
  Google DeepMind\\
  United Kingdom\\
  \texttt{thomaslampe@google.com} \\
  \And
  Abbas Abdolmaleki\\
  Google DeepMind\\
  United Kingdom\\
  \texttt{aabdolmaleki@google.com} \\
  \And
  Jost Tobias Springenberg\\
  Google DeepMind\\
  United Kingdom\\
  \texttt{springenberg@google.com} \\
  \And
  Martin Riedmiller\\
  Google DeepMind\\
  United Kingdom\\
  \texttt{riedmiller@google.com} \\
}

\usepackage{todonotes}
\usepackage{xspace}
\usepackage{soul}
\usepackage{multirow}
\usepackage{caption}
\usepackage{subcaption}
\usepackage{tcolorbox}
\usepackage{graphicx}
\usepackage[T1]{fontenc}
\usepackage{lmodern}
\usepackage{wrapfig}
\usepackage{xcolor}

\definecolor{customgray}{RGB}{220, 220, 220}

\newtcolorbox{graybox}{
  colback=gray!10, 
  colframe=gray!50, 
  boxrule=0.5pt, 
  arc=4pt, 
  left=1mm, 
  right=1mm, 
  top=1mm, 
  bottom=1mm 
}
\newtcolorbox{titlegraybox}[1]{
  colback=gray!10, 
  colframe=gray!50, 
  boxrule=0.5pt, 
  arc=4pt, 
  left=1mm, 
  right=1mm, 
  top=1mm, 
  bottom=1mm, 
  title=#1
}

\newcommand{\skill}[1]{\texttt{<#1>}}

\begin{document}
\maketitle
\def\thefootnote{*}\footnotetext{Equal contribution.}


\begin{abstract}
We propose an agent architecture that automates parts of the common reinforcement learning experiment workflow, to enable automated mastery of control domains for embodied agents.
To do so, it leverages a VLM to perform some of the capabilities normally required of a human \scientist, including the monitoring and analysis of experiment progress, the proposition of new tasks based on past successes and failures of the agent, decomposing tasks into a sequence of subtasks (skills), and retrieval of the skill to execute -- enabling our system to build automated curricula for learning.
We believe this is one of the first proposals for a system that leverages a VLM throughout the full experiment cycle of reinforcement learning.
We provide a first prototype of this system, and examine the feasibility of current models and techniques for the desired level of automation.
For this, we use a standard Gemini model, without additional fine-tuning,
to provide a curriculum of skills to a language-conditioned Actor-Critic algorithm,
in order to steer data collection so as to aid learning new skills.
Data collected in this way is shown to be useful for learning and iteratively improving control policies in a robotics domain.
Additional examination of the ability of the system ability to build a growing library of skills, and to judge the progress of the training of those skills, also shows promising results, suggesting that the proposed architecture provides a potential recipe for fully automated mastery of tasks and domains for embodied agents.
\end{abstract}



\section{Introduction}

\begin{figure}
    \centering
    \includegraphics[width=\textwidth]{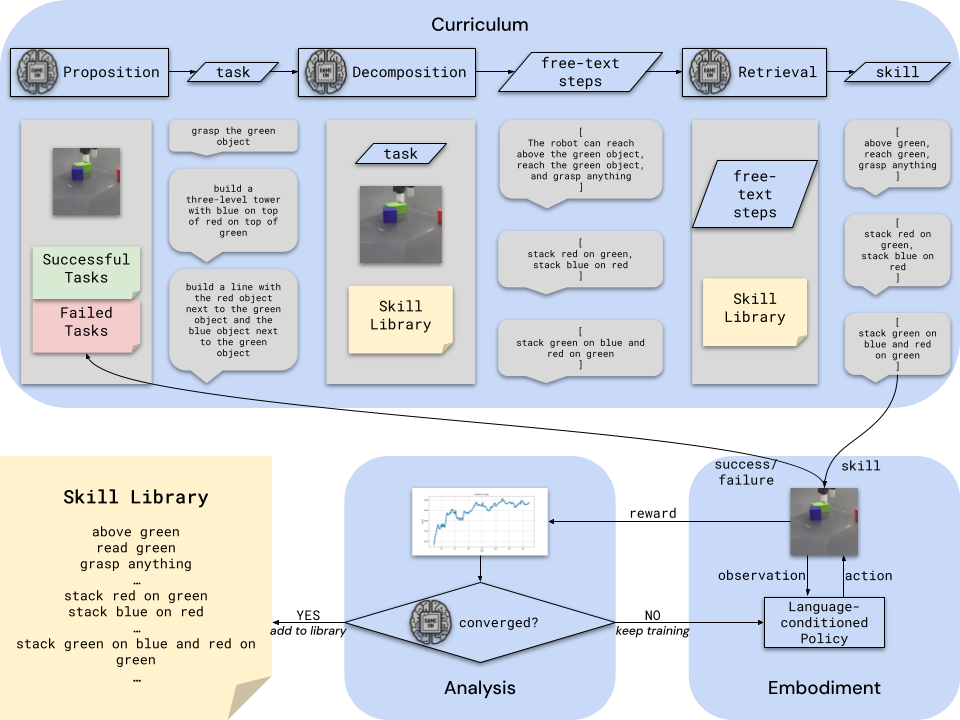}
    \caption{
    Illustration of the system architecture and the interaction of its components.
    The curriculum module generates free-text propositions, decomposes them into free-text steps, and tries to map those onto fixed-text skills from a library.
    We note in the current implementation if the retrieval step (map free-text steps onto fixed-text skills) failed then the plan will be discarded, as we do not have access to a reward model for generating arbitrary reward signals from skill captions; once such a reward model becomes available the failed retrieval should signal the training of a new skill.
    The generated skill sequence is executed by a text-conditioned policy of the embodiment module and unrolled into an episode, which is used to improve the policy.
    Performance during policy training is evaluated by the analysis module, which judges whether training has converged and skills should be added to the library.
    }
    \label{fig:programflow}
\end{figure}

Recent progress on leveraging large (vision) language models (VLMs/LLMs) for reinforcement learning and robotics has demonstrated their usefulness for learning robot policies~\citep{pmlr-v229-zitkovich23a,reed2022generalist,springenberg2024offline} as well as for high-level reasoning~\citep{sermanet2023robovqa,chen2024spatialvlm},
and they have also aided research into automated generation of reward functions for policy learning~\citep{yu2023language,ma2024eureka}. 

In doing so, LLMs have reduced the amount of domain-specific knowledge that an RL researcher would normally need to provide. 
Yet there are still many steps within the experiment workflow of training policies via reinforcement learning (RL) that currently require human intervention;
such as deciding when an experiment has concluded or building a curriculum of tasks~\citep{narvekar2020curriculum,klink2022curriculum} to facilitate the learning of a target task. 
While some work exists in the literature that attempts to automate some of these steps (e.g. automated training and evaluation of standard machine learning tasks~\cite{huang2023benchmarking} or automated curriculum building~\cite{wang2023voyager}
within the community of automated machine learning), these more automated systems usually consider the individual steps in isolation, using models specifically trained to automate a single step.

In this work, we set out to give a first sketch how a large vision language model (VLM) could be used to automate most of the missing capabilities that would be required to automate a reinforcement learning (RL) experiment.
We propose a system architecture that
uses a VLM for automating most parts of the reinforcement learning experiment loop (with the current exception of not providing the reward signal), and trains a growing set of motor skills to increasing mastery of desired domains.
This architecture integrates several of the capabilities traditionally required from a human experimenter:
\begin{itemize}
    \item The proposition of new tasks to perform/learn, given a set of already-known tasks.
    \item The decomposition of higher-level tasks into sequences of low-level skills; paired with retrieving the actual skill the robot possesses.
    \item Judging whether training of a set of skills has concluded, and a new round of data collection should be started for subsequent reinforcement learning.
\end{itemize}

To implement our agent, we focus on examining the suitability of currently available VLMs and prompting techniques for the intended level of automation; rather than attempting to expand their capabilities.
We therefore limit the scope of this implementation to only some of the components of the proposed system.
Notably, we do not automate the stopping of the experiment and the gradual additions of skills to the system yet, and instead present a post-hoc evaluation for both to mimic what their effect would have been.
In addition, due to the unavailability of a robust model that can automatically generate reward functions for arbitrary task, we do not automate the addition of arbitrary new skills and limit our evaluation to a domain with known rewards.

In our prototype, all of the reasoning capabilities are driven by a single, general purpose, and publicly available VLM~\footnote{We use Gemini 1.5- Pro~\citep{team2023gemini} in all experiments.}, and are achieved zero-shot via prompting techniques.
Data generated under this high-level VLM's supervision is then used offline to improve a separate 'low-level' policy, which is trained specifically to output actions to control a robot, and is task-conditioned on language instructions from the high-level system.
To showcase the usefulness of our approach, we train a policy to perform multiple manipulation tasks on a simulated robot.
We show that the VLM-guided exploration produces richer data diversity, which in turn improves performance during successive iterations of policy self-improvement via fine-tuning.
In addition, we show that the same VLM can provide experiment supervision by judging the point at which an experiment should be considered to have converged.
Lastly, we illustrate that if we provide the VLM with growing sets of skills from different stages of the learning process -- as judged by the VLM itself -- it can produce reasonable decompositions for each stage and guide learning of progressively more complicated skills.


\section{Related Work}
\label{sec:citations}

\subsection{LLM-based Virtual Agents}

Following the significant improvements in performance and capabilities of LLMs,
the field of LLM-based agent has seen a surge of recent interest.
Using an LLM as a general-purpose controller, recent work has attempted to replace 
components or capabilities that used to require different pieces of software, models or human researchers by using outputs generated by prompting large language models.

Among these, there are, for example, works that propose general strategies to obtain enhanced inference from agentic LLMs by the use of chain-of-thought reasoning~\cite{wei2022chain},
self-consistency~\cite{wang2022self},
self-reflexion~\cite{shinn2023reflexion},
ReACT chains and tool use~\cite{yao2022react}.

More relevant to our work are the increasing number of LLM-empowered agents proposed to automate science and engineering tasks.
For example, LLM-based software engineer agents are now being designed to assist software development, leveraging the greatly improved coding capabilities of these models. This includes work that utilizes language models to enhance various aspects of software development such as assistive AI pair-programming for interactive notebooks or algorithmic reasoning within competitive programming~\cite{wang2023execution,yin2023natural,shi2024can}. Some recent work goes even further, e.g.
the SWE-Agent~\cite{yang2024sweagent} explores performing end-to-end software engineering with LLM-based agents where a custom agent-computer interface is built to facilitate the agent to navigate repositories, edit code and execute programs.

On the side of automating scientific research,
LLM-based agents have been proposed to perform the work of researchers:
this includes generating novel research directions~\cite{wang2023scimon},
reading through relevant literature to gather information~\cite{lala2023paperqa},
automate discovery of scientific knowledge~\cite{kramer2023automated}, 
come up with hypothesis and revise it based on experimental evidence~\cite{ma2024llm}.
In the specific field of automating machine learning research,
there are studies that use LLMs to help hyper-parameter tuning of machine learning models~\cite{zhang2023automl,zhang2023mlcopilot},
as well as work that gives the LLM-agent capabilities to interact with computer files and execute code; thus conducting machine learning experimentation in a more integrated fashion~\cite{huang2023benchmarking}.

In this work, using a vision language model to both monitor the progress of a machine learning experiment and to examine the resulting performance to influence later experiments is one of the aspects that we focus on.
Different to existing work, 
instead of automating experimentation in purely virtual domains,
we perform experiments with an embodied, robotic, agent in this work for automating RL research and automating domain mastery.

\subsection{LLM/VLM-based Embodied Agents}
Above we have discussed works that utilize LLMs to accomplish pure virtual tasks.
There are, however, also LLM-empowered methods that are designed to assist embodied agents.

For example, in the Minecraft domain, there is work on using  large-scale foundation VLMs to learn behavior policies within the video pre-training (VPT) line of work~\citep{baker2022video,lifshitz2024steve,wang2023jarvis}.
More closely related to our work is the open-ended Voyager agent~\citep{wang2023voyager}.
In particular, in Voyager, GPT-4~\citep{achiam2023gpt} acts as the backbone,
proposing tasks for it to accomplish and writing code to help it achieve goals.
It maintains a skill library which keeps track of LLM-generated and self-verified executable code interfacing with the Minecraft environment through JavaScript API; while in our case the stored skills are learned parameterized low-level control policies rather than code.
They employ an auto-curriculum proposed by LLMs to enable the agent to perform open-ended exploration; we adopt a similar mechanism, but use it to facilitate automatic domain mastery via RL, although we limit our prototype application to one robotic domain with a restricted set of skills that we can easily evaluate. 



In the robotics domain,
CaP~\citep{liang2023code} is closely related to Voyager in the code-writing aspect.
It leverages LLMs to write policy code using perception and control APIs to control robots.
While its follow-up work PromptBook~\citep{arenas2023how} provides further improvements and guidance in prompting LLMs to write code for low-level manipulation control primitives,
the high-level reasoning capability of LLMs is not highlighted in these works. 
Utilizing LLM-based reasoning for robotics tasks was pioneered by SayCan~\citep{brohan2023can} which uses language models to decompose a given high-level task into sub-tasks,
in which the decoding for decomposition is constrained by the availability of the robot skills and weighed by the affordance of skills under a current scene.
In their work, instructions or high-level tasks are provided by human operators rather than suggested by the LLM itself, which restricts the usability of their method for more general and open-ended purposes such as exploration or automatic mastery of domains, as done in this work. Our work is complementary in that we do not restrict the suggestion and decomposition performed by the high-level LLM; but do restrict ourselves to a fixed set of executable low-level skills for which rewards can be computed.
Likewise, \citet{di2023towards} also uses LLMs to decompose tasks into sub-goals, and uses those as instructions for a language-conditioned policy.
Similar to SayCan, the tasks in their work are explicitly provided by a user.
Furthermore, it requires a separate VLM to obtain text descriptions from visual observations, as well as fine-tuning of an LLM to their specific domain.
This is in contrast to our work where a native multimodal Gemini~\citep{team2023gemini} model is used without the need for finetuning.
On the task-proposing front, there are several works that focus on simulated domains: \citet{wang2023gensim} propose to leverage LLMs for both proposing tasks and generating simulation code for the proposed tasks, while \citet{xian2023towards} further sketch a system that also includes components like code generation for reward.
Since we are interested in applying our proposed system directly in the real world, these methods would need further adaptation, e.g. by adding automatic reward modeling methods that do not require access to the simulator state.

A perhaps most closely related LLM-assisted agent in the control/robotics domain to our work is AutoRT~\citep{ahn2024autort}.
They adopt an LLM-based approach to orchestrate a fleet of robots to collect diverse data,
where tasks are proposed on-the-fly by LLMs.
While the open-ended task proposition is very similar to the Voyager paradigm and our work,
there are several notable differences to our approach.
First, AutoRT is purely aimed at data collection without any active learning of policies 'in the loop' wheras we set out to specifically automate the process of automatically steering the learning process of a low level RL learner.
Secondly, since their proposed tasks are at skill-level already,
there is no decomposition component in their agent architecture and particularly no sequencing of skills.
The authors also only measure the diversity of the collected data, but do not validate the quality of the data empirically by conducting any type of policy training with it.
In contrast, we do use the data collected by our proposed system to perform self-improvement and bootstrap a more capable policy.

In order to eventually execute and evaluate arbitrary tasks, plans and subgoals proposed by an automated \scientist, one key capability would be to get a reward function from the language caption of the task. As discussed above the current work does not yet include such a step.
While we do not investigate this aspect in this work (we restrict our self-improvement experiments on training skills which the reward function is known), we do note that there is a growing body of works that utilize LLMs to write reward code for desired behavior \cite{yu2023language,liang2024learning,ma2024eureka} and VLMs as general reward models or success detectors \cite{rocamonde2023vision,du2023vision}, which are candidates for integration into our system once reliable enough for use in reinforcement learning domains.






\section{System Architecture}
\label{sec:architecture}

We study the setting of an embodied agent with access to a workspace containing objects that it can interact with.
We propose a VLM-based agent architecture that can enable automatic mastery of the environment. By mastery we here mean that we expect the agent to be capable of accomplishing any meaningful task -- for which we can measure success by a given set of reward functions -- with any object in the environment by the end of the learning process,
and by automatic we mean that no human researcher is required to come up with a decomposition of tasks or a curriculum for learning the tasks in a specific order during the learning process and that no researcher is needed to monitor the training progress.
Our proposed agent architecture to fulfill this goal consists of  the following modules:

\begin{itemize}
    \item
The curriculum module, which performs high-level reasoning to guide the learning process with auto-curriculum.
More specifically,
it is in charge of task proposition, task decomposition, skill retrieval, 
and keeps a record of past successful and failed episodes.
    \item
The embodiment module, which maintains a skill library consisting of the skills available to the embodiment. It will execute skills assigned by the curriculum in the environment, save episode data and report back success or failure. Finally it will trigger a low-level 'Actor-Critic' RL algorithm for learning (or improving) skills from the collected data.
    \item
The analysis module, which monitors the training progress of skills, reports learning status and adds converged ones to the skill library of the corresponding embodiment.
\end{itemize}


\subsection{The Curriculum Module}
\label{sec:curriculum}
This module generates an auto-curriculum to guide automatic mastery of domains.
Each of its components (task proposition, task decomposition, skill retrieval) is realized by prompting the Gemini model.

In the following, we describe each component's prompts conceptually.
For concrete prompt designs used in our prototype implementation, see Appendix \ref{appendix:prompt_design}.

\paragraph{Task proposition.}
We first prompt the VLM to propose a new high-level task in free-form text; for the agent to accomplish given a current image observation and past success and failures.
The VLM is prompted to output tasks that are novel and diverse while not being too far from the current capability of the agent.
The proposition prompt is heavily inspired by the prompt used in Voyager \cite{wang2023voyager}, 
consisting of a description of the domain, the request for a proposal and matching reasoning leading to it, as well as a set of constraints and requirements.
This fixed prompt and a number of exemplars are followed by a current image of the domain, and a growing list of successfully and unsuccessfully completed proposals from the same experiment. We refer to the appendix for a description of the instruction and how exemplars images and success detection are included in the prompt.


\begin{graybox}
\texttt{proposition\_prompt = ( \\
\hspace*{2em} proposition\_instruction + proposition\_exemplars +
\\
\hspace*{2em} current\_image\_observation + successful\_trials + failed\_trials
\\)
\\
proposition\_reasoning, proposed\_task = VLM(proposition\_prompt)}
\end{graybox}

\paragraph{Task decomposition.}
Given a free-text high-level task proposition, the list of available skills and a current image observation, we then prompt the VLM to decompose the task into a list of sub-goals/sub-tasks, again described as fee-form text without any restriction.
The decomposition prompt contains a general instruction and several exemplars, to which we concatenate the free-form description of the proposed task from the previous task, the encoding of a current image of the domain, and the skills currently available (note that these skills are naturally limited to those we can evaluate a reward for in our current implementation, and thus are fixed text strings rather than free-form text).
We note that although the fixed-text skills are provided to the decomposition prompt, the decomposition is not instructed to output steps using those fixed-text skills only, this is intended such that it should come up with steps that are necessary,
which may or may not be available (and thus we could use the steps as a proposal for learning a new skill that should be added to our library); the effect of the availability of the returned free-text skills will be discussed further in the retrieval section.


\begin{graybox}
\texttt{decomposition\_prompt = (
\\\hspace*{2em}decomposition\_instruction + decomposition\_exemplars 
\\)
\\decomposition\_reasoning, [subtask\_0, subtask\_1, ..., subtask\_n] = VLM(
\\\hspace*{2em}decomposition\_prompt + proposed\_task + current\_image\_observation +
\\\hspace*{2em}skill\_library
\\)}
\end{graybox}

\paragraph{Skill retrieval.}
Given each decomposed free-text sub-task and the list of available fixed-text skills, we can finally retrieve the most semantically similar skill -- from the available skills -- to accomplish the desired sub-task.
Note that several previous works perform retrieval using the embeddings of the language instructions,
whereas we formulate it as a direct question answering (QA) task in text, which we find to be more robust.
The retrieval prompt consists of a general instruction that states the retrieval request and the constraint of not rephrasing the retrieved skill name but picking only among the available skills, followed by an exemplar, then appended with the current list of available fixed-text skills and the free-text decomposition step to map. The result of this step then is a plan of a sequence of skills from the library that can be executed by the agent.

\begin{graybox}
\texttt{plan = []
\\for query\_subtask in [subtask\_0, subtask\_1, ..., subtask\_n]:
\\\hspace*{2em}retrieval\_prompt = (
\\\hspace*{4em}retrieval\_instruction + retrieval\_exemplars +
\\\hspace*{4em}query\_subtask + skill\_library
\\\hspace*{2em})
\\\hspace*{2em}retrieval\_reasoning, retrieved\_skill = VLM(retrieval\_prompt)
\\\hspace*{2em}plan.append(retrieved\_skill)
}
\end{graybox}

In addition, if the retrieval of a step in the decomposition sequence fails,
this can serve as a signal that a new skill is required in the policy.
Such cases can then be included in the next round of policy learning by the embodiment module.
We note, however, that due to the lack of a system to generate rewards for arbitrary skills, in this work we manually perform the choice of skills to add; which also means that in the current implementation, the decomposition plan will be discarded if the retrieval of any of its steps failed.

\subsection{The Embodiment Module}
\label{sec:embodiment}
After obtaining a decomposed list of subgoals, the curriculum will communicate the task to the corresponding embodiment to execute.
After executing a sequence of skills, it judges the success of the sequence, i.e. whether the goal that led to the decomposition, has been achieved, and reports the result back to the Curriculum module.
While determining the success of a sequence relies on pre-defined reward functions in our prototype system (see section \ref{sec:system_interaction} for details),
it could, in theory, also draw upon LLM-based reward functions.

The module also collects all executed episodes in a dataset.
Once a stopping criterion has been reached -- classically pre-defined as a certain number of episodes, but potentially also triggered by the Analysis module in section \ref{sec:analysis} -- a new policy learning iteration is launched with this dataset to fine-tune the previous policy; any offline policy learning algorithm could be used in this step and we refer to the next section for our specific exemplary choice.
For increased data efficiency, all of the episodes are re-labeled with the rewards of all of the skills currently known to the agent, including those newly added by the curriculum module described in section \ref{sec:curriculum}.

\subsection{The Analysis Module}
\label{sec:analysis}
Finally, this module examines the learning progress of skills by few-shot prompting.
The prompt prefix is formatted as:

\begin{graybox}
\texttt{You are an assistant for monitoring the progress of reinforcement learning experiments.
Tell whether the learning has converged or not given the plot of the curve of the accumulated reward it obtains. 
Give a concise reasoning of your examination result and the answer in YES or NO only.
}
\end{graybox}

Each exemplar is given in the format of:

\begin{graybox}
\texttt{\textbf{Reward curve:} \{reward\_plot\_image\}. \\
\textbf{Reasoning:} \{reasoning\}. \\
\textbf{A:} \{has\_converged\}.
}
\end{graybox}

where the reward curve plot for each exemplar is plugged into 
\texttt{\{reward\_plot\_image\}},  
and the exemplar reasoning and answer are placed into
\texttt{\{reasoning\}} and
\texttt{\{has\_converged\}}.


For all skills for a certain embodiment,
the analysis module will periodically go through the learning curves of each of them.
Those that are judged as converged will be added to the available skills of that embodiment (and by extension, become available to the curriculum module for decomposition) and its training will be terminated.

We note that this imposes some constraints on the reward functions that can be used: they need to be normalized, and the episode duration of evaluation runs needs to be known, in order to allow meaningful scaling of the curves for analysis.


\section{System Realization}
\label{sec:implementation}

In order to explore the feasibility of the system, we implement its components, and apply them to a simulated robotic manipulation task.

\subsection{Module Interaction}
\label{sec:system_interaction}

\newcommand{\gaiaagent}[0]{curriculum module\xspace}
\newcommand{\policyagent}[0]{embodiment module\xspace}
\newcommand{\xmagent}[0]{analysis module\xspace}



%

The \gaiaagent periodically retrieves images from the environment, and includes them into the goal proposal prompt.
The goal is then decomposed into steps and skill captions are retrieved.
If any of the steps cannot be mapped to a known skill during retrieval, the plan is discarded, and the process repeated.
If all steps are retrieved, the skill sequence is sent to the \policyagent, which uses them to condition a text-conitioned learned policy; we use the perceiver-actor-critic (PAC) algorithm \citep{springenberg2024offline} to learn and represent such policies.
The program flow is controlled by the \gaiaagent: after each decomposition, all of the potentially multiple \policyagent instances perform a fixed number of episode rollouts, with the skill being changed at fixed (pre-defined) intervals.
We acknowledge that this approach only applies for quasi-static domains like the object arrangement tasks considered here.
For more dynamic domains, it is necessary to also condition the model to return a duration for each skill, or to continuously query it as to whether to switch skills at a given point in time.

At the end of each rollout, the \policyagent reports whether the plan was successfully executed. Success here is defined as observing a reward $>0.5$ for each executed skill, and $>0.95$ for the last skill in the sequence;
all skills in the proposed sequence must be completed to qualify as success.
The \gaiaagent includes these success reports into its list of successful and unsuccessful plans, for use in subsequent prompts.

\begin{figure}
    \centering
    \includegraphics[width=\textwidth]{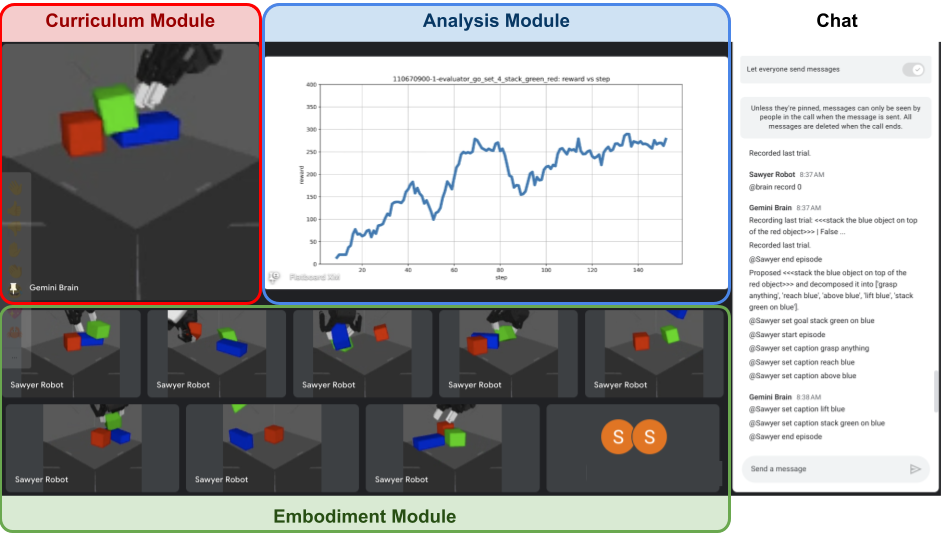}
    \caption{
    Screenshot of a Google Meet session hosting the agent when performing a multi-robot simulation experiment, with boxes annotating the different modules.
    Note that the \gaiaagent mirrors the image stream of one of the {\policyagent}s it currently attends to.
    }
    \label{fig:meet}
\end{figure}

We use a chat-based interface between these modules, similar to that used by \citet{sermanet2023robovqa}.
This allows easily connecting them in a natural interface, which also facilitates human introspection and intervention during testing.
Modules simply join a Google Meet session, and interact with each other via chat messages, as well as streaming image and video data through it.
Messages can be broadcast, enabling a single high-level VLM to control the skills of multiple low-level policies at the same time, thus increasing compute efficiency in the face of otherwise expensive queries to the VLM.
The setup is illustrated in Figure \ref{fig:meet}.

The \xmagent is used outside of the experiment loop in this prototype.
Rather than actually stopping the experiment, we run it after the experiment has concluded, so that we can evaluate whether the termination point chosen by it was indeed the point of convergence.

\subsection{Policy Training}
For the low-level control policy, we employ a Perceiver-Actor-Critic (PAC) model~\citep{springenberg2024offline}.
Such a model has been shown to be trainable via offline reinforcement learning, can be text conditioned, and is able to utilize non-expert (exploration) data that our agent will generate.
This in turn allows us to additionally relabel all data with multiple reward functions, and thus reuse a small amount of data to train all desired skills.

In PAC, skills can be represented by either conditioning the policy on language, on goal images, or a mixture thereof. Here, we purely opt for language, as this allows us to directly communicate the high-level system's skill proposals to the low-level policy.

\subsection{Prompting}

The high-level system is represented by a standard Gemini 1.5 Pro model~\citep{team2023gemini}.
To design the prompts for the Gemini model, we use the publicly available
OneTwo Python library~\citep{onetwo}.
OneTwo is a model-agnostic layer that abstracts aspects such as injecting components into a VLM prompt template, and extracting pre-defined fields from the model's response.

Each component's prompt contains a small number of exemplars which were hand-designed and include image data from previous experiments.
This includes 2 each for for proposal and decomposition, 1 for retrieval, and 6 for analysis.
It is also worth noting that none of the proposal exemplars contain a scene with three objects, unlike in the domain we apply it to, in order to not bias the responses.
All exemplars used are provided in Appendix \ref{appendix:prompt_exemplars}


\section{Experimental Results}
\label{sec:result}

\subsection{Benchmark}

To evaluate the benefits of our approach, we consider a robotic block stacking task, previously described in \citet{bousmalis2023robocat}.
In this task, three colored objects in a basket need to be arranged into a desired configuration by a 7-DoF Franka Panda robot fitted with a Robotiq 2F-85 parallel gripper.
The domain is implemented in the MuJoCo simulator~\citep{todorov2012mujoco}.
This task was chosen since it provides combinatorial complexity, which lends itself to building up more complex skills, yet is also narrow enough to allow automatic evaluation and manual reward design.

More specifically, our expectation with this domain is for the Gemini-aided auto-curriculum to be able to lead the agent to automatically discover and learn the object configurations such as tower and pyramid, which were previously manually designed by human researchers.

\subsection{Auto-curriculum-based Exploration}

To examine the ability of the system to perform task proposal and decomposition, we first train a PAC model to perform a number of simple base skills for the \gaiaagent to utilize.
Note that the framework also allows for the agent to learn from scratch, but here as a proof of concept, we start with a base set of skills to allow for faster learning iterations.
We use a pre-existing dataset of approximately 1M episodes collected from a single-task RL experiment, where an MPO agent~\citep{abdolmaleki2018maximum} was trained to perform the different permutations of stacking a single object on top of another.
The data is re-labeled with reward functions corresponding to a set of basic skills,
including opening and closing the gripper, reaching an object, lifting an object, holding one object over another, and stacking them. For a full list see Appendix \ref{appendix:skill_rewards}.
We then train a PAC model with 140M parameters for 1.5M steps, after which performance has stabilized for all skills.

We then use this fixed policy to perform Gemini-driven data collection, following the approach described in Section \ref{sec:system_interaction}.
As this data is intended for further self-improvement training, we roughly follow the CHEF approach~\citep{lampe2023mastering} of performing a hyperparameter search to aid diversity.
However, we do not vary the parameters of the slow PAC training, but instead explore different settings for the \gaiaagent.
Firstly, we vary the sampling temperature of the VLM, using both $0.0$ and $0.3$.
Secondly, we perform collection runs with different sets of skills made available to the agent: either all of the skills including the composite \skill{stack A on B}, or only simpler ones up to \skill{hold A above B}.

In each run, the \gaiaagent controls 10 simulator instances in order to parallelize data collection.
Each skill proposal and decomposition sequence is also executed 5 times per robot to reduce querying load on the VLM.
Decomposed plans are executed open-loop, in the sense that each skill in the sequence is maintained for a fixed duration of 20 seconds before switching to the next one.
In this manner, we collect a set of 25k robot episodes in total.

\subsubsection{Data Diversity}

\newcommand{\robocatdata}[0]{\texttt{pretraining}\xspace}
\newcommand{\gaiadata}[0]{\texttt{self-improvement}\xspace}

First, we compare the dataset used for pre-training the PAC policy (which we refer to as \robocatdata set) with the new dataset collected by our method (\gaiadata set), using a distance metric similar to \citet{brohan2023can}.
We do this separately for camera images and proprioception data (i.e. joint angles).
For camera images, we use a CoCa image embedding~\citep{yu2022coca}; for proprioception, we use the non-embedded observations, and normalize them first along each dimension and then overall for unit norm.
Then we measure the relative L2 distance of these representations to each other, as well as the distance of each to their respective cluster in a k-means clustering with 5 clusters (where the clusters were learned on the \robocatdata set).
Table \ref{tab:embedding_distance} highlights how data in the collected \gaiaagent set appears to be more spread out.
The diversity in vision and proprioception data can be taken to be directly beneficial for self-improvement.

\begin{table}[]
    \centering
    \begin{tabular}{|c|cc|cc|cc|} \hline
             & \multicolumn{2}{c|}{Vision} & \multicolumn{2}{c|}{Proprioception} \\
             & L2 & Cluster  & L2 & Cluster \\ \hline
\robocatdata & 0.571 & 0.380 & 0.815 & 0.369 \\
\gaiadata    & 0.625 & 0.473 & 0.901 & 0.551  \\ \hline
    \end{tabular}
    \caption{Diversity metrics for original and collected datasets, per modality.}
    \label{tab:embedding_distance}
\end{table}

Separately, we also compare the diversity in the embeddings of the language instruction of the skills executed throughout the episodes.
We pass these through the embedding available via an older, text-only Gemini model, and contrast the diversity of the \robocatdata set with the combined one used for fine-tuning.
We observe an L2 distance of 0.287 and cluster distance of 0.097 for the \robocatdata set, vs. 0.555 L2 and 0.732 cluster for the combined set.
This matches our expectations, given that the original diversity was low (with only 6 permutation of the form "stack A on B"), the \gaiadata set not only executes more diverse skills, but also multiple per episode.



When inspecting the \gaiaagent data visually,
it also becomes apparent that it generates more complex object arrangements.
For instance, we find multiple attempts to build a tower, as well as pyramid-like structures -- the latter of which result e.g. from failed tower building attempts, as the model does not propose pyramid-building itself.

It is also worth noting that the proposals generated by the model are in fact fairly focused, and mostly cover plans such as \skill{stack A on B and stack X on Y}, as well as \skill{put A next to B}, for only 27 unique proposals. However, the decomposition module expands these into 102 unique skill sequences.
For instance, decomposing the task of building a red-blue-green tower at different times results in two plans with the same outcome but different skill sequences:
\begin{graybox}
\skill{reach blue}, \skill{lift blue}, \skill{above green>}, \skill{open gripper}, \\ \skill{reach red}, \skill{lift red}, \skill{above blue}, \skill{open gripper}
\end{graybox}
\begin{graybox}
\skill{reach blue}, \skill{lift blue}, \skill{reach green}, \skill{above green}, \skill{open gripper}, \\ \skill{reach red}, \skill{lift red}, \skill{reach blue}, \skill{above blue}, \skill{open gripper>}
\end{graybox}

Additional examples, both successful and failed, are provided in Appendix
\ref{appendix:response_examples}.

\subsubsection{Self-Improvement}


In addition to quantifying the quality of the collected data, we also use it to perform a round of self-improvement of the pretrained PAC policy.
For this, we introduce three new skill into the set learned by the model that were not available in the skill library given to the \gaiaagent for data collection: arranging the three objects into a pyramid shape, arranging them in an inverted pyramid, and arranging them to form a tower of three objects.
We manually chose these for being the same as previously used as benchmark by \citet{bousmalis2023robocat};
while the curriculum module frequently suggests tower building, it does not suggest the pyramid tasks during exploration.
The rewards for these tasks are defined in Appendix \ref{appendix:skill_rewards}.

Data is relabeled with these new rewards in addition to the existing ones.
We then fine-tune the PAC model with two datasets: once using only the original \robocatdata dataset, and once using the combined \robocatdata and \gaiadata sets.
In the latter case, given that the \gaiadata set is much smaller than the \robocatdata set, we up-sample it so that both datasets contribute 50\% of each training batch.
We also up-sample the newly added pyramid-building skill, to in turn account for 50\% of the data from each dataset.

\begin{figure}
    \centering
    \begin{subfigure}{.32\textwidth}
    \includegraphics[width=\textwidth]{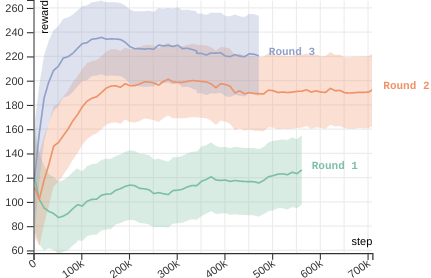}
    \caption{All tasks}
    \end{subfigure}
    \begin{subfigure}{.32\textwidth}
    \includegraphics[width=\textwidth]{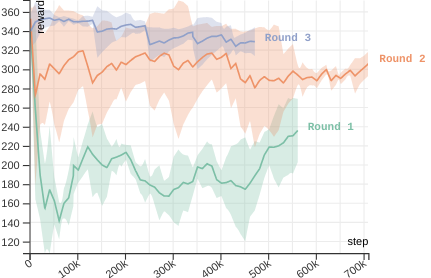}
    \caption{\texttt{reach}}
    \end{subfigure}
    \begin{subfigure}{.32\textwidth}
    \includegraphics[width=\textwidth]{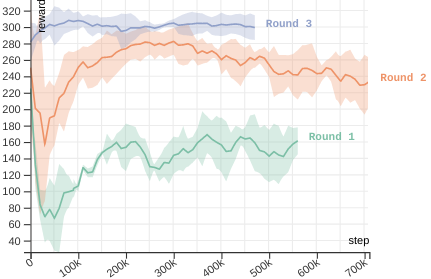}
    \caption{\texttt{lift}}
    \end{subfigure}
    \begin{subfigure}{.32\textwidth}
    \includegraphics[width=\textwidth]{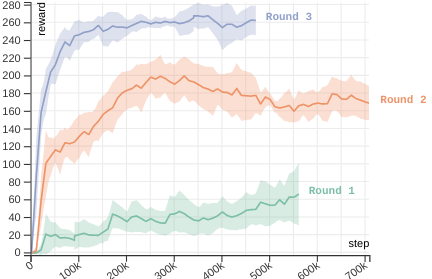}
    \caption{\texttt{stack}}
    \end{subfigure}
    \begin{subfigure}{.32\textwidth}
    \includegraphics[width=\textwidth]{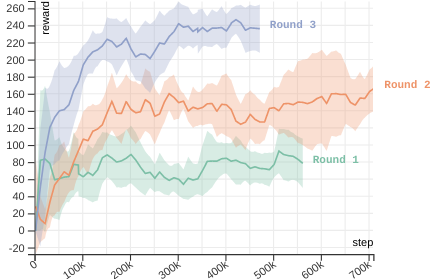}
    \caption{\texttt{pyramid}}
    \end{subfigure}
    \begin{subfigure}{.32\textwidth}
    \includegraphics[width=\textwidth]{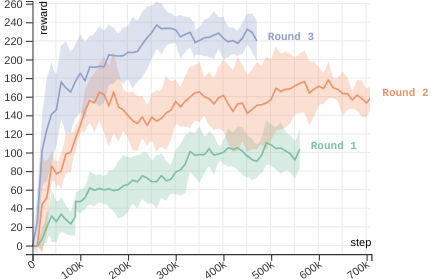}
    \caption{\texttt{inverted pyramid}}
    \end{subfigure}
    \caption{Training curves for PAC self-improvement. Adding the \gaiadata set  (red) consistently outperforms using only the \robocatdata set (green), both on the average of all tasks (top left), and on all individual task families.
    Adding then a third set of data collected with the newly added pyramid skills, produces even better performance (blue).
    }
    \label{fig:pac_improvement}
\end{figure}

Figure \ref{fig:pac_improvement} compares the performance of these datasets on a selection of skills.
As is evident from these results, not only does the added data allow the model to learn the "pyramid" skills, but it also leads to better performance on the base skills.
It is worth noting that none of the policy learns to perform tower building; this is due to the low success rate of the pretrained PAC policy when sequencing multiple skills in order to attempt stacking (since this leads to visiting states that are not represented in the original \robocatdata data).
Failed tower building does often lead to "accidental" creation of pyramids however, which explains the better performance on those tasks.
We therefore point out that it seems sensible to separate the proposal of new tasks to learn from the proposal of tasks used during data collection.

Finally, using the skills resulting from learning on the combined datasets we perform one additional iteration: we subsequently collect 15k more episodes with the newly trained pyramid building skills added to the skill library; and thus available to the curriculum module.
We again up-sample data so as to weight all three data sources equally.
When using this data to fine-tune the PAC policy once more, performance for these skills increases substantially, as also seen in Figure \ref{fig:pac_improvement}.




\subsection{VLM-based Performance Analysis}
\label{sec:performance_analysis}

During the initial PAC policy training, we trained the model for approximately 1.5M learner steps.
After running that long, we observe a degeneration of performance, particularly for "simpler" skills, which can be attributed to overfitting.
Normally, a human RL \scientist would employ early stopping to avoid such effects, and stop the experiment once the learning curves for all skills appear to have converged.
Here, we use Gemini to judge the convergence state of the experiment post-hoc after the training has concluded and run for an extended number of steps, in order to determine the point at which the model would have proposed early stopping.

All learning curves are scaled to a maximum reward of $400$, which is known since rewards are clipped to $[0; 1]$ and evaluation episodes do not exceed $400$ steps.
The analysis model is not otherwise informed regarding the expected total reward of each skill.

\begin{figure}
    \centering
    \includegraphics[width=\textwidth]{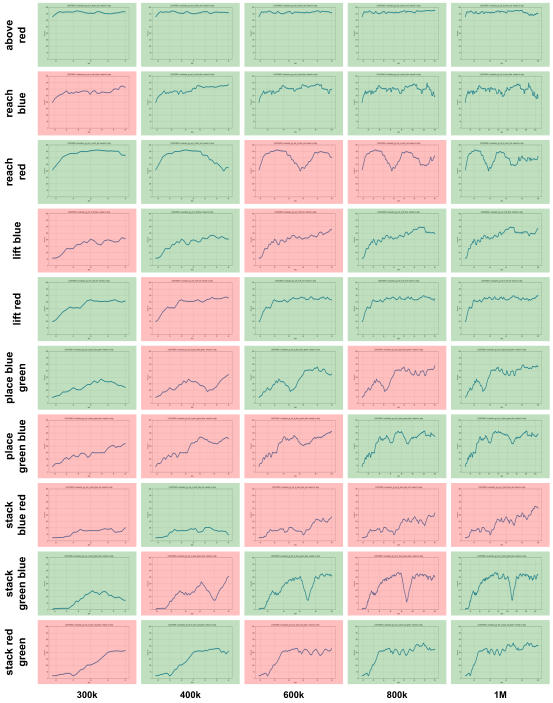}
    \caption{Example evaluation curves provided to the VLM, at different numbers of learning steps during training. Color coding denotes whether a curve was judged as converged (green) or not yet converged (red).
    }
    \label{fig:convergence_matrix}
\end{figure}

\begin{figure}
    \centering
    \includegraphics[width=\textwidth]{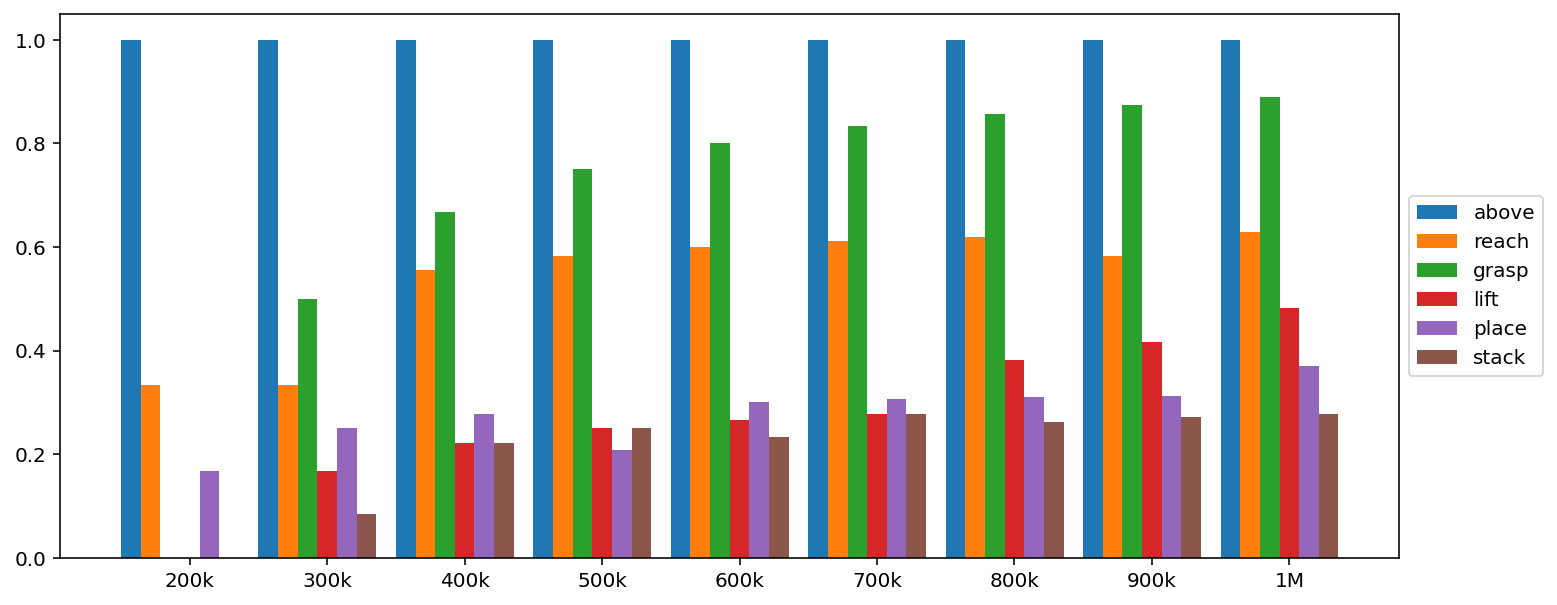}
    \caption{Cumulative percentage of curves judged as converged by the analysis module, per task family.}
    \label{fig:cumulative_analysis_success}
\end{figure}

Figure \ref{fig:convergence_matrix} illustrates a selection of these judgments.
While these judgments are not fully stable, and false ones do occur, the VLM judges an increasing number of skills as converged as training progresses.
Evident errors occur mostly when judging early plateaus in the learning progress (e.g. \skill{place blue on green} at 300k steps or \skill{stack red on green} at 400k steps) -- a limitation that would similarly affect a human practitioner if not aware of the expected final reward.
Other unstable classifications involve irregular curves such as those of \skill{lift red}.
Overall, the ratings reflect both the increasing performance of the skills over time, as well as their relative difficulty, as illustrated in Figure \ref{fig:cumulative_analysis_success}, where easier skills can be seen to be judged as converged from early on, while harder ones only get judged as such later on average.

\subsection{Progressively Adding Skills}
\label{sec:progressive_skills}

\begin{table}[h]
    \centering
    \includegraphics[width=\textwidth]{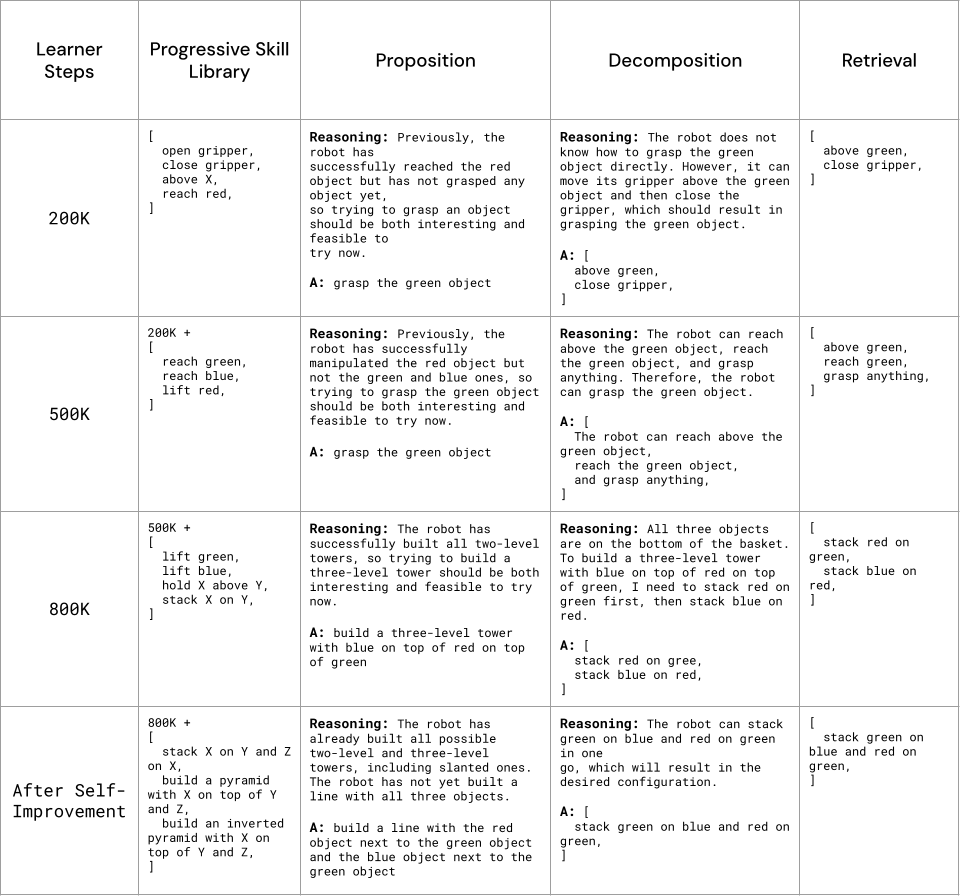}
    \caption{Example propositions, decompositions and retrievals provided by the curriculum module when provided growing skill libraries from different experiment stages; all propositions and decompositions in this table take in the same image observation containing three objects (red, green, blue) shown in Appendix \ref{appendix:progressive_skills}. Note that the \texttt{Reasoning} part of the proposition response is cut short to contain only the last segment, the full response are reported in Appendix \ref{appendix:progressive_skills}.}
    \label{fig:progressive_skills}
\end{table}

A second purpose of the analysis module lies in determining which skills are trained sufficiently to be used for decomposition.
In this work, we first trained the PAC policy to convergence, before starting curriculum-driven data collection.
But generally, these two processes can be performed concurrently.
In order to illustrate the curriculum module's ability to work with a growing set of skills, we therefore examine some of the plans generated when using those skills judged as converged in section \ref{sec:performance_analysis} at certain points in time.

We examine the proposals and decompositions at four points of the experiment: with those skills judged successful after 200k, 500k and 800k learner updates in the first PAC training experiment, as well as the entire set of skills added for self-improvement.

An overview of the model responses is provided in Table \ref{fig:progressive_skills}.
For more detailed outputs of the model, including the reasoning provided by the model for each response, see Appendix \ref{appendix:progressive_skills}.

We see that after both 200k and 500k steps, the proposition yields the same simple goal.
But while after 200k steps the system has to use the most basic skills for decomposition, it employs the more reliable \skill{reach green} and \skill{grasp anything} skills at 500k steps.
At 800k steps, when all skills are available, it generates more complex propositions, and directly uses the higher-level stacking skills.
And with the fine-tuning skills included, the model attempts to arrange the objects into a line, which resolves into building a tower -- i.e. a vertical line.


\section{Discussion and Future Work}
\label{sec:conclusion}

We have outlined an agent architecture for reinforcement learning that uses a VLM to perform various capabilities normally expected of a human \scientist in Section \ref{sec:architecture}.
These capabilities would allow automating the training process of the agent beyond current capabilities, and let an embodied agent autonomously acquire an ever-growing set of skills with ever-increasing mastery.

We implemented and evaluated a first prototype of such a system in Section \ref{sec:implementation}, including the functionalities of proposing new tasks for exploration, decomposing them into skill sequences, and analyzing the progress of the learning experiment.
For this first proof-of-concept system, we simplified several of the components and their interaction.
This was done both to limit the scope of this study, but also in order to focus on determining whether state-of-the-art methods and models are able to perform the required capabilities -- particularly when used zero-shot, without costly fine-tuning of the VLM.

The prototype system showed the ability to automatically collect diverse data, which was successfully used to perform self-improvement of the control policy, and to learn new skills not learnable with a more narrow starting set (section \ref{sec:result}).
The curriculum also displayed the ability to adapt the complexity of its task propositions and plans to the complexity of the available skills.

Going forward, we intend to reduce the simplifications made for the prototype system and strive for full automation, with several natural next steps outlined in the following.

In our prototype implementation, the analysis of learning progress was performed post-hoc, to illustrate the quality of analysis at different stages.
While the quality of judgments of experiment convergence was not fully reliable and suffered from the same uncertainty at plateaus that a human does, it did show potential to make correct judgments when aggregated over time.
In addition, the successful application of the system to self-improvement of the policy means that even if prematurely terminating the training of skills at plateaus, training of these skills can continue in subsequent self-improvement rounds.
In the future, we would therefore seek to integrate it directly into the automation and allow it to stop the experiment.

In the future, we also plan to include LLM-based reward functions into the architecture, once techniques are mature enough.
While currently it is too easy for RL agents to exploit false reward detections, recent advances such as Eureka~\citep{ma2024eureka} promise zero- or few-shot LLM-based rewards without the need to train specialized models.
This provides a natural next \scientist's capability to integrate: to have the system automatically add its proposals as new skills.
It will also allow adding the one functionality of the \gaiaagent we left out thus far: to automatically add proposed actions as new skills -- which requires the PAC training to be able to label datasets with rewards matching those proposals.

Related to this, we currently simply discard decomposed sequences if any of the steps cannot be mapped to a skill known by the policy.
But encountering an unknown step provides a strong signal that there is a skill missing, and a natural next step is to include it in the next PAC training cycle as a base skill.
However, doing so would again require a universal reward module, which is not presently available.

During evaluation of the PAC policies, we observed that skills are often not sequenced correctly, e.g. after completing skill \skill{stack red on blue}, the policy may not perform \skill{stack green on red} successfully, even though it can perform it in isolation.
This may be either attributed to incomplete separation between skills, or as insufficient data coverage; generally, the base dataset would never have observed the terminal state of one skill as the initial state of another.
This can lead to otherwise sensible plans generated by the decomposition module to not achieve task success, which in turn causes the \skill{stack X on Y and Z on X} skills to never achieve non-zero performance after the PAC finetuning.
We hypothesize that this may be rectified through repeated self-improvement, as the skill sequencing would generate more diverse data, and/or by reducing the weight of the narrowly biased \robocatdata set.
However, such extended data collection was not feasible in this work.

During episode unrolls, we executed the skill sequence open-loop, maintaining each skill for a fixed amount of time.
This was justified by the fact that the domain is largely static, but we do note that for more dynamic domains, it would be necessary to have the decomposition provide a duration for each skill, or determine the switching point dynamically.
The latter is infeasible for Gemini-sized models due to limited inference speed, but is expected to become possible in the future, as smaller yet equally capable models become available.



\clearpage
\acknowledgments{We thank Ksenia Konyushkova for suggestions on prompting, Pierre Sermanet for laying the foundations of the technical infrastructure, and Jonas Adler for general discussions of the research field.}


\bibliography{example}  

\begin{thebibliography}{46}
\providecommand{\natexlab}[1]{#1}
\providecommand{\url}[1]{\texttt{#1}}
\expandafter\ifx\csname urlstyle\endcsname\relax
  \providecommand{\doi}[1]{doi: #1}\else
  \providecommand{\doi}{doi: \begingroup \urlstyle{rm}\Url}\fi

\bibitem[Zitkovich et~al.(2023)Zitkovich, Yu, Xu, Xu, Xiao, Xia, Wu, Wohlhart,
  Welker, Wahid, Vuong, Vanhoucke, Tran, Soricut, Singh, Singh, Sermanet,
  Sanketi, Salazar, Ryoo, Reymann, Rao, Pertsch, Mordatch, Michalewski, Lu,
  Levine, Lee, Lee, Leal, Kuang, Kalashnikov, Julian, Joshi, Irpan, Ichter,
  Hsu, Herzog, Hausman, Gopalakrishnan, Fu, Florence, Finn, Dubey, Driess,
  Ding, Choromanski, Chen, Chebotar, Carbajal, Brown, Brohan, Arenas, and
  Han]{pmlr-v229-zitkovich23a}
B.~Zitkovich, T.~Yu, S.~Xu, P.~Xu, T.~Xiao, F.~Xia, J.~Wu, P.~Wohlhart,
  S.~Welker, A.~Wahid, Q.~Vuong, V.~Vanhoucke, H.~Tran, R.~Soricut, A.~Singh,
  J.~Singh, P.~Sermanet, P.~R. Sanketi, G.~Salazar, M.~S. Ryoo, K.~Reymann,
  K.~Rao, K.~Pertsch, I.~Mordatch, H.~Michalewski, Y.~Lu, S.~Levine, L.~Lee,
  T.-W.~E. Lee, I.~Leal, Y.~Kuang, D.~Kalashnikov, R.~Julian, N.~J. Joshi,
  A.~Irpan, B.~Ichter, J.~Hsu, A.~Herzog, K.~Hausman, K.~Gopalakrishnan, C.~Fu,
  P.~Florence, C.~Finn, K.~A. Dubey, D.~Driess, T.~Ding, K.~M. Choromanski,
  X.~Chen, Y.~Chebotar, J.~Carbajal, N.~Brown, A.~Brohan, M.~G. Arenas, and
  K.~Han.
\newblock Rt-2: Vision-language-action models transfer web knowledge to robotic
  control.
\newblock In J.~Tan, M.~Toussaint, and K.~Darvish, editors, \emph{Proceedings
  of The 7th Conference on Robot Learning}, volume 229 of \emph{Proceedings of
  Machine Learning Research}, pages 2165--2183. PMLR, 06--09 Nov 2023.
\newblock URL \url{https://proceedings.mlr.press/v229/zitkovich23a.html}.

\bibitem[Reed et~al.(2022)Reed, Zolna, Parisotto, Colmenarejo, Novikov,
  Barth-Maron, Gimenez, Sulsky, Kay, Springenberg, Eccles, Bruce, Razavi,
  Edwards, Heess, Chen, Hadsell, Vinyals, Bordbar, and
  de~Freitas]{reed2022generalist}
S.~Reed, K.~Zolna, E.~Parisotto, S.~G. Colmenarejo, A.~Novikov, G.~Barth-Maron,
  M.~Gimenez, Y.~Sulsky, J.~Kay, J.~T. Springenberg, T.~Eccles, J.~Bruce,
  A.~Razavi, A.~Edwards, N.~Heess, Y.~Chen, R.~Hadsell, O.~Vinyals, M.~Bordbar,
  and N.~de~Freitas.
\newblock A generalist agent, 2022.

\bibitem[Springenberg et~al.(2024)Springenberg, Abdolmaleki, Zhang, Groth,
  Bloesch, Lampe, Brakel, Bechtle, Kapturowski, Hafner, Heess, and
  Riedmiller]{springenberg2024offline}
J.~T. Springenberg, A.~Abdolmaleki, J.~Zhang, O.~Groth, M.~Bloesch, T.~Lampe,
  P.~Brakel, S.~Bechtle, S.~Kapturowski, R.~Hafner, N.~Heess, and
  M.~Riedmiller.
\newblock Offline actor-critic reinforcement learning scales to large models,
  2024.

\bibitem[Sermanet et~al.(2023)Sermanet, Ding, Zhao, Xia, Dwibedi,
  Gopalakrishnan, Chan, Dulac-Arnold, Maddineni, Joshi, Florence, Han, Baruch,
  Lu, Mirchandani, Xu, Sanketi, Hausman, Shafran, Ichter, and
  Cao]{sermanet2023robovqa}
P.~Sermanet, T.~Ding, J.~Zhao, F.~Xia, D.~Dwibedi, K.~Gopalakrishnan, C.~Chan,
  G.~Dulac-Arnold, S.~Maddineni, N.~J. Joshi, P.~Florence, W.~Han, R.~Baruch,
  Y.~Lu, S.~Mirchandani, P.~Xu, P.~Sanketi, K.~Hausman, I.~Shafran, B.~Ichter,
  and Y.~Cao.
\newblock Robo{VQA}: Multimodal long-horizon reasoning for robotics, 2023.

\bibitem[Chen et~al.(2024)Chen, Xu, Kirmani, Ichter, Driess, Florence, Sadigh,
  Guibas, and Xia]{chen2024spatialvlm}
B.~Chen, Z.~Xu, S.~Kirmani, B.~Ichter, D.~Driess, P.~Florence, D.~Sadigh,
  L.~Guibas, and F.~Xia.
\newblock Spatialvlm: Endowing vision-language models with spatial reasoning
  capabilities.
\newblock \emph{arXiv preprint arXiv:2401.12168}, 2024.

\bibitem[Yu et~al.(2023)Yu, Gileadi, Fu, Kirmani, Lee, Arenas, Chiang, Erez,
  Hasenclever, Humplik, et~al.]{yu2023language}
W.~Yu, N.~Gileadi, C.~Fu, S.~Kirmani, K.-H. Lee, M.~G. Arenas, H.-T.~L. Chiang,
  T.~Erez, L.~Hasenclever, J.~Humplik, et~al.
\newblock Language to rewards for robotic skill synthesis.
\newblock In \emph{Conference on Robot Learning}, pages 374--404. PMLR, 2023.

\bibitem[Ma et~al.(2024)Ma, Liang, Wang, Huang, Bastani, Jayaraman, Zhu, Fan,
  and Anandkumar]{ma2024eureka}
Y.~J. Ma, W.~Liang, G.~Wang, D.-A. Huang, O.~Bastani, D.~Jayaraman, Y.~Zhu,
  L.~Fan, and A.~Anandkumar.
\newblock Eureka: Human-level reward design via coding large language models,
  2024.

\bibitem[Narvekar et~al.(2020)Narvekar, Peng, Leonetti, Sinapov, Taylor, and
  Stone]{narvekar2020curriculum}
S.~Narvekar, B.~Peng, M.~Leonetti, J.~Sinapov, M.~E. Taylor, and P.~Stone.
\newblock Curriculum learning for reinforcement learning domains: A framework
  and survey.
\newblock \emph{Journal of Machine Learning Research}, 21\penalty0
  (181):\penalty0 1--50, 2020.

\bibitem[Klink et~al.(2022)Klink, Yang, D’Eramo, Peters, and
  Pajarinen]{klink2022curriculum}
P.~Klink, H.~Yang, C.~D’Eramo, J.~Peters, and J.~Pajarinen.
\newblock Curriculum reinforcement learning via constrained optimal transport.
\newblock In \emph{International Conference on Machine Learning}, pages
  11341--11358. PMLR, 2022.

\bibitem[Huang et~al.(2023)Huang, Vora, Liang, and
  Leskovec]{huang2023benchmarking}
Q.~Huang, J.~Vora, P.~Liang, and J.~Leskovec.
\newblock Benchmarking large language models as {AI} research agents.
\newblock In \emph{NeurIPS 2023 Foundation Models for Decision Making
  Workshop}, 2023.

\bibitem[Wang et~al.(2023)Wang, Xie, Jiang, Mandlekar, Xiao, Zhu, Fan, and
  Anandkumar]{wang2023voyager}
G.~Wang, Y.~Xie, Y.~Jiang, A.~Mandlekar, C.~Xiao, Y.~Zhu, L.~Fan, and
  A.~Anandkumar.
\newblock Voyager: An open-ended embodied agent with large language models,
  2023.

\bibitem[{Gemini Team Google}(2023)]{team2023gemini}
{Gemini Team Google}.
\newblock Gemini: A family of highly capable multimodal models.
\newblock \emph{arXiv preprint arXiv:2312.11805}, 2023.

\bibitem[Wei et~al.(2022)Wei, Wang, Schuurmans, Bosma, Xia, Chi, Le, Zhou,
  et~al.]{wei2022chain}
J.~Wei, X.~Wang, D.~Schuurmans, M.~Bosma, F.~Xia, E.~Chi, Q.~V. Le, D.~Zhou,
  et~al.
\newblock Chain-of-thought prompting elicits reasoning in large language
  models.
\newblock \emph{Advances in neural information processing systems},
  35:\penalty0 24824--24837, 2022.

\bibitem[Wang et~al.(2022)Wang, Wei, Schuurmans, Le, Chi, Narang, Chowdhery,
  and Zhou]{wang2022self}
X.~Wang, J.~Wei, D.~Schuurmans, Q.~Le, E.~Chi, S.~Narang, A.~Chowdhery, and
  D.~Zhou.
\newblock Self-consistency improves chain of thought reasoning in language
  models.
\newblock \emph{arXiv preprint arXiv:2203.11171}, 2022.

\bibitem[Shinn et~al.(2023)Shinn, Labash, and Gopinath]{shinn2023reflexion}
N.~Shinn, B.~Labash, and A.~Gopinath.
\newblock Reflexion: an autonomous agent with dynamic memory and
  self-reflection.
\newblock \emph{arXiv preprint arXiv:2303.11366}, 2023.

\bibitem[Yao et~al.(2022)Yao, Zhao, Yu, Du, Shafran, Narasimhan, and
  Cao]{yao2022react}
S.~Yao, J.~Zhao, D.~Yu, N.~Du, I.~Shafran, K.~R. Narasimhan, and Y.~Cao.
\newblock React: Synergizing reasoning and acting in language models.
\newblock In \emph{The Eleventh International Conference on Learning
  Representations}, 2022.

\bibitem[Wang et~al.(2023)Wang, Zhou, Fried, and Neubig]{wang2023execution}
Z.~Wang, S.~Zhou, D.~Fried, and G.~Neubig.
\newblock Execution-based evaluation for open-domain code generation.
\newblock In \emph{The 2023 Conference on Empirical Methods in Natural Language
  Processing}, 2023.

\bibitem[Yin et~al.(2023)Yin, Li, Xiao, Rao, Wen, Shi, Howland, Bailey,
  Catasta, Michalewski, et~al.]{yin2023natural}
P.~Yin, W.-D. Li, K.~Xiao, A.~K. Rao, Y.~Wen, K.~Shi, J.~Howland, P.~Bailey,
  M.~Catasta, H.~Michalewski, et~al.
\newblock Natural language to code generation in interactive data science
  notebooks.
\newblock In \emph{The 61st Annual Meeting Of The Association For Computational
  Linguistics}, 2023.

\bibitem[Shi et~al.(2024)Shi, Tang, Narasimhan, and Yao]{shi2024can}
Q.~Shi, M.~Tang, K.~Narasimhan, and S.~Yao.
\newblock Can language models solve olympiad programming?
\newblock \emph{arXiv preprint arXiv:2404.10952}, 2024.

\bibitem[Yang et~al.(2024)Yang, Jimenez, Wettig, Lieret, Yao, Narasimhan, and
  Press]{yang2024sweagent}
J.~Yang, C.~E. Jimenez, A.~Wettig, K.~Lieret, S.~Yao, K.~Narasimhan, and
  O.~Press.
\newblock {SWE}-agent: Agent-computer interfaces enable automated software
  engineering, 2024.

\bibitem[Wang et~al.(2023)Wang, Downey, Ji, and Hope]{wang2023scimon}
Q.~Wang, D.~Downey, H.~Ji, and T.~Hope.
\newblock Scimon: Scientific inspiration machines optimized for novelty.
\newblock \emph{arXiv preprint arXiv:2305.14259}, 2023.

\bibitem[L{\'a}la et~al.(2023)L{\'a}la, ODonoghue, Shtedritski, Cox, Rodriques,
  and White]{lala2023paperqa}
J.~L{\'a}la, O.~ODonoghue, A.~Shtedritski, S.~Cox, S.~G. Rodriques, and
  A.~White.
\newblock Paperqa: Retrieval-augmented generative agent for scientific
  research.
\newblock 2023.

\bibitem[Kramer et~al.(2023)Kramer, Cerrato, D{\v{z}}eroski, and
  King]{kramer2023automated}
S.~Kramer, M.~Cerrato, S.~D{\v{z}}eroski, and R.~King.
\newblock Automated scientific discovery: from equation discovery to autonomous
  discovery systems.
\newblock \emph{arXiv preprint arXiv:2305.02251}, 2023.

\bibitem[Ma et~al.(2024)Ma, Wang, Guo, Sun, Tenenbaum, Rus, Gan, and
  Matusik]{ma2024llm}
P.~Ma, T.-H. Wang, M.~Guo, Z.~Sun, J.~B. Tenenbaum, D.~Rus, C.~Gan, and
  W.~Matusik.
\newblock Llm and simulation as bilevel optimizers: A new paradigm to advance
  physical scientific discovery.
\newblock \emph{arXiv preprint arXiv:2405.09783}, 2024.

\bibitem[Zhang et~al.(2023{\natexlab{a}})Zhang, Gong, Wu, Liu, and
  Zhou]{zhang2023automl}
S.~Zhang, C.~Gong, L.~Wu, X.~Liu, and M.~Zhou.
\newblock Automl-gpt: Automatic machine learning with gpt.
\newblock \emph{arXiv preprint arXiv:2305.02499}, 2023{\natexlab{a}}.

\bibitem[Zhang et~al.(2023{\natexlab{b}})Zhang, Zhang, Ren, Li, and
  Yang]{zhang2023mlcopilot}
L.~Zhang, Y.~Zhang, K.~Ren, D.~Li, and Y.~Yang.
\newblock {ML}copilot: Unleashing the power of large language models in solving
  machine learning tasks.
\newblock \emph{arXiv preprint arXiv:2304.14979}, 2023{\natexlab{b}}.

\bibitem[Baker et~al.(2022)Baker, Akkaya, Zhokov, Huizinga, Tang, Ecoffet,
  Houghton, Sampedro, and Clune]{baker2022video}
B.~Baker, I.~Akkaya, P.~Zhokov, J.~Huizinga, J.~Tang, A.~Ecoffet, B.~Houghton,
  R.~Sampedro, and J.~Clune.
\newblock Video pretraining (vpt): Learning to act by watching unlabeled online
  videos.
\newblock \emph{Advances in Neural Information Processing Systems},
  35:\penalty0 24639--24654, 2022.

\bibitem[Lifshitz et~al.(2024)Lifshitz, Paster, Chan, Ba, and
  McIlraith]{lifshitz2024steve}
S.~Lifshitz, K.~Paster, H.~Chan, J.~Ba, and S.~McIlraith.
\newblock Steve-1: A generative model for text-to-behavior in minecraft.
\newblock \emph{Advances in Neural Information Processing Systems}, 36, 2024.

\bibitem[Wang et~al.(2023)Wang, Cai, Liu, Ma, and Liang]{wang2023jarvis}
Z.~Wang, S.~Cai, A.~Liu, X.~Ma, and Y.~Liang.
\newblock Jarvis-1: Open-world multi-task agents with memory-augmented
  multimodal language models.
\newblock In \emph{Second Agent Learning in Open-Endedness Workshop}, 2023.

\bibitem[Achiam et~al.(2023)Achiam, Adler, Agarwal, Ahmad, Akkaya, Aleman,
  Almeida, Altenschmidt, Altman, Anadkat, et~al.]{achiam2023gpt}
J.~Achiam, S.~Adler, S.~Agarwal, L.~Ahmad, I.~Akkaya, F.~L. Aleman, D.~Almeida,
  J.~Altenschmidt, S.~Altman, S.~Anadkat, et~al.
\newblock Gpt-4 technical report.
\newblock \emph{arXiv preprint arXiv:2303.08774}, 2023.

\bibitem[Liang et~al.(2023)Liang, Huang, Xia, Xu, Hausman, Ichter, Florence,
  and Zeng]{liang2023code}
J.~Liang, W.~Huang, F.~Xia, P.~Xu, K.~Hausman, B.~Ichter, P.~Florence, and
  A.~Zeng.
\newblock Code as policies: Language model programs for embodied control.
\newblock In \emph{2023 IEEE International Conference on Robotics and
  Automation (ICRA)}, pages 9493--9500. IEEE, 2023.

\bibitem[Arenas et~al.(2023)Arenas, Xiao, Singh, Jain, Ren, Vuong, Varley,
  Herzog, Leal, Kirmani, Sadigh, Sindhwani, Rao, Liang, and
  Zeng]{arenas2023how}
M.~G. Arenas, T.~Xiao, S.~Singh, V.~Jain, A.~Z. Ren, Q.~Vuong, J.~Varley,
  A.~Herzog, I.~Leal, S.~Kirmani, D.~Sadigh, V.~Sindhwani, K.~Rao, J.~Liang,
  and A.~Zeng.
\newblock How to prompt your robot: A promptbook for manipulation skills with
  code as policies.
\newblock In \emph{Towards Generalist Robots: Learning Paradigms for Scalable
  Skill Acquisition @ CoRL2023}, 2023.
\newblock URL \url{https://openreview.net/forum?id=1aRNtmy5zX}.

\bibitem[Brohan et~al.(2023)Brohan, Chebotar, Finn, Hausman, Herzog, Ho, Ibarz,
  Irpan, Jang, Julian, et~al.]{brohan2023can}
A.~Brohan, Y.~Chebotar, C.~Finn, K.~Hausman, A.~Herzog, D.~Ho, J.~Ibarz,
  A.~Irpan, E.~Jang, R.~Julian, et~al.
\newblock Do as {I} can, not as {I} say: Grounding language in robotic
  affordances.
\newblock In \emph{Conference on robot learning}, pages 287--318. PMLR, 2023.

\bibitem[Di~Palo et~al.(2023)Di~Palo, Byravan, Hasenclever, Wulfmeier, Heess,
  and Riedmiller]{di2023towards}
N.~Di~Palo, A.~Byravan, L.~Hasenclever, M.~Wulfmeier, N.~Heess, and
  M.~Riedmiller.
\newblock Towards a unified agent with foundation models.
\newblock In \emph{Workshop on Reincarnating Reinforcement Learning at ICLR
  2023}, 2023.

\bibitem[Wang et~al.(2023)Wang, Ling, Yuan, Shridhar, Bao, Qin, Wang, Xu, and
  Wang]{wang2023gensim}
L.~Wang, Y.~Ling, Z.~Yuan, M.~Shridhar, C.~Bao, Y.~Qin, B.~Wang, H.~Xu, and
  X.~Wang.
\newblock Gensim: Generating robotic simulation tasks via large language
  models.
\newblock In \emph{The Twelfth International Conference on Learning
  Representations}, 2023.

\bibitem[Xian et~al.(2023)Xian, Gervet, Xu, Qiao, Wang, and
  Wang]{xian2023towards}
Z.~Xian, T.~Gervet, Z.~Xu, Y.-L. Qiao, T.-H. Wang, and Y.~Wang.
\newblock Towards generalist robots: A promising paradigm via generative
  simulation.
\newblock \emph{arXiv preprint arXiv:2305.10455}, 2023.

\bibitem[Ahn et~al.(2024)Ahn, Dwibedi, Finn, Arenas, Gopalakrishnan, Hausman,
  Ichter, Irpan, Joshi, Julian, Kirmani, Leal, Lee, Levine, Lu, Leal,
  Maddineni, Rao, Sadigh, Sanketi, Sermanet, Vuong, Welker, Xia, Xiao, Xu, Xu,
  and Xu]{ahn2024autort}
M.~Ahn, D.~Dwibedi, C.~Finn, M.~G. Arenas, K.~Gopalakrishnan, K.~Hausman,
  B.~Ichter, A.~Irpan, N.~Joshi, R.~Julian, S.~Kirmani, I.~Leal, E.~Lee,
  S.~Levine, Y.~Lu, I.~Leal, S.~Maddineni, K.~Rao, D.~Sadigh, P.~Sanketi,
  P.~Sermanet, Q.~Vuong, S.~Welker, F.~Xia, T.~Xiao, P.~Xu, S.~Xu, and Z.~Xu.
\newblock Auto{RT}: Embodied foundation models for large scale orchestration of
  robotic agents, 2024.

\bibitem[Liang et~al.(2024)Liang, Xia, Yu, Zeng, Arenas, Attarian, Bauza,
  Bennice, Bewley, Dostmohamed, et~al.]{liang2024learning}
J.~Liang, F.~Xia, W.~Yu, A.~Zeng, M.~G. Arenas, M.~Attarian, M.~Bauza,
  M.~Bennice, A.~Bewley, A.~Dostmohamed, et~al.
\newblock Learning to learn faster from human feedback with language model
  predictive control.
\newblock \emph{arXiv preprint arXiv:2402.11450}, 2024.

\bibitem[Rocamonde et~al.(2023)Rocamonde, Montesinos, Nava, Perez, and
  Lindner]{rocamonde2023vision}
J.~Rocamonde, V.~Montesinos, E.~Nava, E.~Perez, and D.~Lindner.
\newblock Vision-language models are zero-shot reward models for reinforcement
  learning.
\newblock In \emph{The Twelfth International Conference on Learning
  Representations}, 2023.

\bibitem[Du et~al.(2023)Du, Konyushkova, Denil, Raju, Landon, Hill, de~Freitas,
  and Cabi]{du2023vision}
Y.~Du, K.~Konyushkova, M.~Denil, A.~Raju, J.~Landon, F.~Hill, N.~de~Freitas,
  and S.~Cabi.
\newblock Vision-language models as success detectors.
\newblock In \emph{Conference on Lifelong Learning Agents}, pages 120--136.
  PMLR, 2023.

\bibitem[one()]{onetwo}
One{T}wo.
\newblock \url{https://github.com/google-deepmind/onetwo}.
\newblock Accessed: 2024-05-29.

\bibitem[Bousmalis et~al.(2023)Bousmalis, Vezzani, Rao, Devin, Lee, Bauza,
  Davchev, Zhou, Gupta, Raju, Laurens, Fantacci, Dalibard, Zambelli, Martins,
  Pevceviciute, Blokzijl, Denil, Batchelor, Lampe, Parisotto, Żołna, Reed,
  Colmenarejo, Scholz, Abdolmaleki, Groth, Regli, Sushkov, Rothörl, Chen,
  Aytar, Barker, Ortiz, Riedmiller, Springenberg, Hadsell, Nori, and
  Heess]{bousmalis2023robocat}
K.~Bousmalis, G.~Vezzani, D.~Rao, C.~Devin, A.~X. Lee, M.~Bauza, T.~Davchev,
  Y.~Zhou, A.~Gupta, A.~Raju, A.~Laurens, C.~Fantacci, V.~Dalibard,
  M.~Zambelli, M.~Martins, R.~Pevceviciute, M.~Blokzijl, M.~Denil,
  N.~Batchelor, T.~Lampe, E.~Parisotto, K.~Żołna, S.~Reed, S.~G. Colmenarejo,
  J.~Scholz, A.~Abdolmaleki, O.~Groth, J.-B. Regli, O.~Sushkov, T.~Rothörl,
  J.~E. Chen, Y.~Aytar, D.~Barker, J.~Ortiz, M.~Riedmiller, J.~T. Springenberg,
  R.~Hadsell, F.~Nori, and N.~Heess.
\newblock Robocat: A self-improving generalist agent for robotic manipulation,
  2023.

\bibitem[Todorov et~al.(2012)Todorov, Erez, and Tassa]{todorov2012mujoco}
E.~Todorov, T.~Erez, and Y.~Tassa.
\newblock Mu{J}o{C}o: A physics engine for model-based control.
\newblock In \emph{2012 IEEE/RSJ International Conference on Intelligent Robots
  and Systems}, pages 5026--5033, 2012.
\newblock \doi{10.1109/IROS.2012.6386109}.

\bibitem[Abdolmaleki et~al.(2018)Abdolmaleki, Springenberg, Tassa, Munos,
  Heess, and Riedmiller]{abdolmaleki2018maximum}
A.~Abdolmaleki, J.~T. Springenberg, Y.~Tassa, R.~Munos, N.~Heess, and
  M.~Riedmiller.
\newblock Maximum a posteriori policy optimisation, 2018.

\bibitem[Lampe et~al.(2023)Lampe, Abdolmaleki, Bechtle, Huang, Springenberg,
  Bloesch, Groth, Hafner, Hertweck, Neunert, Wulfmeier, Zhang, Nori, Heess, and
  Riedmiller]{lampe2023mastering}
T.~Lampe, A.~Abdolmaleki, S.~Bechtle, S.~H. Huang, J.~T. Springenberg,
  M.~Bloesch, O.~Groth, R.~Hafner, T.~Hertweck, M.~Neunert, M.~Wulfmeier,
  J.~Zhang, F.~Nori, N.~Heess, and M.~Riedmiller.
\newblock Mastering stacking of diverse shapes with large-scale iterative
  reinforcement learning on real robots, 2023.

\bibitem[Yu et~al.(2022)Yu, Wang, Vasudevan, Yeung, Seyedhosseini, and
  Wu]{yu2022coca}
J.~Yu, Z.~Wang, V.~Vasudevan, L.~Yeung, M.~Seyedhosseini, and Y.~Wu.
\newblock Co{C}a: Contrastive captioners are image-text foundation models,
  2022.

\end{thebibliography}


\clearpage
\appendix


\section{Prompt Design}
\label{appendix:prompt_design}

Below are the concrete prompts used in our prototype system.
These contain the static parts of the prompts and the format of exemplars.
For actual exemplars used, see Appendix \ref{appendix:prompt_exemplars}.

\subsection{Task proposition}
 We note that this prompt is heavily inspired by the curriculum prompt in Voyager \cite{wang2023voyager}.

\begin{graybox}
\texttt{You are an assistant for proposing tasks for a robot to perform; the robot has a single arm with a black gripper with two white fingers, it also has a camera looking into the workspace in front of it. \\
Propose the next task for the robot to perform given: an image observation of the current workspace, a list of the completed tasks so far, a list of failed tasks that are too hard. \\
Give a concise reasoning of your proposal, including listing all possible spatial structures achievable by the objects you see, and then give your proposed task; you should propose tasks that form a curriculum such as to help the robot to accomplish as many diverse tasks as possible, build as many different structures as possible, following these criteria: \\
1. The next task should follow a concise format, such as "put [object\_1] next to [object\_2]", "stack [object\_1] on top of [object\_2]" etc. "build a [spatial\_structure] with [object\_1] and [object\_2] on top of [object\_3]",  etc.
It should be a single phrase.
Do not propose multiple tasks at the same time.
Do not mention anything else. \\
2. The next task should not be too hard since the robot may not have learned enough skills to complete it yet. \\
3. The next task should be novel and interesting. The robot should look for different objects to manipulate and different object configurations to achieve. You should not ask the robot to perform the same thing over and over again. \\
4. The robot may sometimes need to repeat some tasks if it needs to collect more data to complete more difficult tasks. Only repeat if necessary. \\
5. The proposed task should ideally be composable of the current skills available to the robot. The robot should look for different objects to manipulate and different object configurations to achieve such that at the end of the curriculum it has "mastered" the whole workspace.
}
\end{graybox}

Each exemplar is given in the format of:

\begin{graybox}
\texttt{\textbf{Image observation of the current workspace:} \{image\_observation\}.
\\\textbf{Completed tasks so far:} \{successful\_trials\}.
\\\textbf{Failed tasks that are too hard:} \{failed\_trials\}.
\\\textbf{Reasoning:} \{reasoning\}.
\\\textbf{A:} \{proposed\_task\}.
}
\end{graybox}

The evaluation content at run time will then be given in the format of:

\begin{graybox}
\texttt{\textbf{Image observation of the current workspace:} \{image\_observation\}.
\\\textbf{Completed tasks so far:} \{successful\_trials\}.
\\\textbf{Failed tasks that are too hard:} \{failed\_trials\}.
}
\end{graybox}

And the model will respond in the following format:

\begin{graybox}
\texttt{\textbf{Reasoning:} \{reasoning\}.
\\\textbf{A:} \{proposed\_task\}.
}
\end{graybox}

\subsection{Task decomposition}

\begin{graybox}
\texttt{You are an assistant for helping a robot completing a given task by decomposing it into a sequence of subtasks; the robot has a single arm with a black gripper with two white fingers, it also has a camera looking into the workspace in front of it. \\
Decompose the given task into subtasks that the robot knows how to perform given: an image observation of the current workspace, a list of the available skills of the robot. \\
Give a concise reasoning of your decomposition and then give your result as a python list of strings, each string contains a decomposed subtask.
}
\end{graybox}

Each exemplar is given in the format of:

\begin{graybox}
\texttt{\textbf{Q}: \{task\}.
\\\textbf{Image observation of the current workspace:} \{image\_observation\}.
\\\textbf{Available skills:} \{available\_skills\}.
\\\textbf{Reasoning:} \{reasoning\}.
\\\textbf{A:} \{decomposed\_task\}.
}
\end{graybox}

The evaluation content at run time will then be given in the format of:

\begin{graybox}
\texttt{\textbf{Q}: \{task\}.
\\\textbf{Image observation of the current workspace:} \{image\_observation\}.
\\\textbf{Available skills:} \{available\_skills\}.
}
\end{graybox}

And the model will respond in the following format:

\begin{graybox}
\texttt{\textbf{Reasoning:} \{reasoning\}.
\\\textbf{A:} \{decomposed\_task\}.
}
\end{graybox}

\subsection{Skill retrieval}

\begin{graybox}
\texttt{You are an assistant for retrieval.
\\Find the most semantically similar entry from a skill library given a query skill description. \\
Pay attention to the object configurations induced by the skill, give a concise reasoning about the result and return the exact entry from the library without rephrasing it.
}
\end{graybox}

Each exemplar is given in the format of:

\begin{graybox}
\texttt{\textbf{Q:} \{query\_skill\}.
\\\textbf{Skill library:} \{available\_skills\}.
\\\texttt{\textbf{Reasoning:} \{reasoning\}.
\\\textbf{A:} \{retrieved\_skill\}.}
}
\end{graybox}

The evaluation content at run time will then be given in the format of:

\begin{graybox}
\texttt{\textbf{Q:} \{query\_skill\}.
\\\textbf{Skill library:} \{available\_skills\}.
}
\end{graybox}

And the model will respond in the following format:

\begin{graybox}
\texttt{\textbf{Reasoning:} \{reasoning\}.
\\\textbf{A:} \{retrieved\_skill\}.
}
\end{graybox}


\section{Prompt Exemplars}
\label{appendix:prompt_exemplars}

This section provides the concrete exemplars that are in the prompt templates described above when querying the model.

\newcommand{\propositionexemplar}[5]{
\begin{titlegraybox}{Proposition exemplar #1}
\texttt{\textbf{Image observation of the current workspace:} }\includegraphics[width=0.15\textwidth]{exemplars/proposition_#1.jpg}
\\\texttt{\textbf{Completed tasks so far:} #2}
\\\texttt{\textbf{Failed tasks that are too hard:} #3} 
\\\texttt{\textbf{Reasoning:} #4} 
\\\texttt{\textbf{A:} #5}
\end{titlegraybox}
}

\newcommand{\decompositionexemplar}[5]{
\begin{titlegraybox}{Decomposition exemplar #1}
\texttt{\textbf{Q:} #2
\\\textbf{Image observation of the current workspace:} }\includegraphics[width=0.15\textwidth]{exemplars/decomposition_#1.jpg}
\\\texttt{\textbf{Available skills:} #3}
\\\texttt{\textbf{Reasoning:} #4} 
\\\texttt{\textbf{A:} #5}
\end{titlegraybox}
}

\newcommand{\retrievalexemplar}[5]{
\begin{titlegraybox}{Retrieval exemplar #1}
\texttt{\textbf{Q:} #2
\\\texttt{\textbf{Skill library:} #3}
\\\texttt{\textbf{Reasoning:} #4} 
\\\texttt{\textbf{A:} #5}
}
\end{titlegraybox}
}

\newcommand{\analysisexemplar}[3]{
\begin{titlegraybox}{Analysis exemplar #1}
\texttt{\textbf{Reward curve:}}\includegraphics[width=0.3\textwidth]{exemplars/analysis_#1.jpg} 
\\\texttt{\textbf{Reasoning:} #2} 
\\\texttt{\textbf{A:} #3}
\end{titlegraybox}
}

\subsection{Proposition}

\propositionexemplar{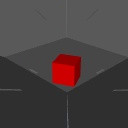}{
    [open gripper, grasp the red object]
}{
    []
}{
    I see one object: red. All possible spatial structures that can be built with it: move the red object to a desired 3D point. Previously, the robot has successfully grasped the red object, so to reach a different spatial structure, you can try move it to a different planar position or lift it up.
}{
    lift the red object up
}

\propositionexemplar{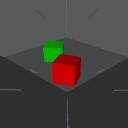}{
    [open gripper, grasp the red object, lift the red object up]
}{
    [stack the green object on top of the red object]
}{
    I see two objects: red, green. All possible spatial structures that can be built with them: a line where the two objects are placed next to each other; two dots where the two objects are apart from each other; a two-level tower with one object on top of another. a two-level slanted tower with one object on top of another but not aligned at the center. Previously, the robot has successfully manipulated the red object but not the green one, and it also has not built any structure with both two objects, so trying to build a two-level tower by stacking the red object on top of the green object should be both interesting and feasible to try now.
}{
    stack the red object on top of the green object
}

\subsection{Task decomposition}

\decompositionexemplar{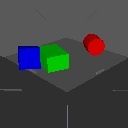}{
    put the red object next to the blue object
}{
    [open gripper, grasp the red object, lift the red object up, stack the red object on top of the blue object]
}{
    All three objects are on the bottom of the basket. The robot is able to stack the red object on top of the blue object but does not know how to put the red one next to the blue one. However, the task might be accomplished by first stacking the red on blue, then lift up the red object and open the gripper, in which case the red object should drop not too far from the blue object.
}{
    [stack the red object on top of the blue object, lift the red object up, open gripper]
}

\decompositionexemplar{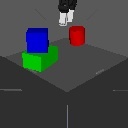}{
    build a three-level tower
}{
    [open gripper, grasp the blue object, stack the blue object on top of the green object, grasp the red object, lift the red object up, stack the red object on top of the blue object]
}{
    The blue object is stacked on top of the green one. To build a three-level tower, I just need to stack the red object on top of the blue object, which I happen to know how to.
}{
    [stack the red object on top of the blue object]
}

\subsection{Retrieval}

\retrievalexemplar{1}{
    stack red on blue
}{
    [open gripper, grasp the red object, lift the red object up, put the red object on top of the blue one]
}{
    The skill in query asks for a object configuration where the red object is on top of the blue one. Except for the 4th one, all other skills in the library concerns with the red object only, while the 4th one not only concerns the red and the blue object but also matches the desired configuration.
}{
   put the red object on top of the blue one
}

\subsection{Analysis}

\analysisexemplar{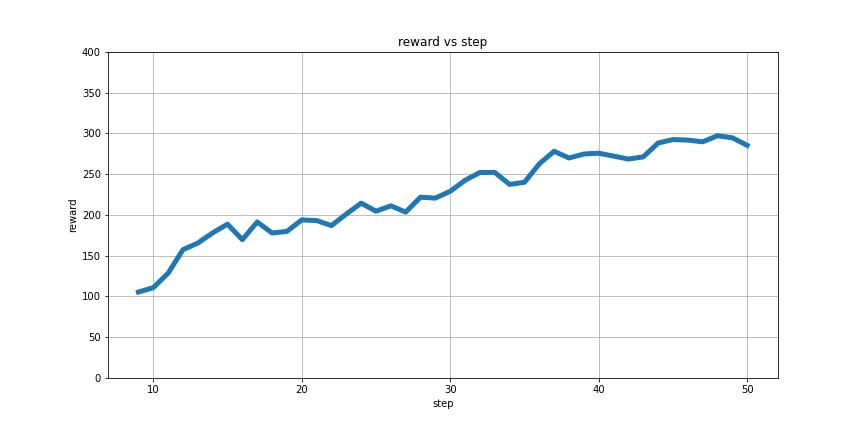}{
    The learning curve is still going up, the learning has not converged yet.
}{
    NO
}
\analysisexemplar{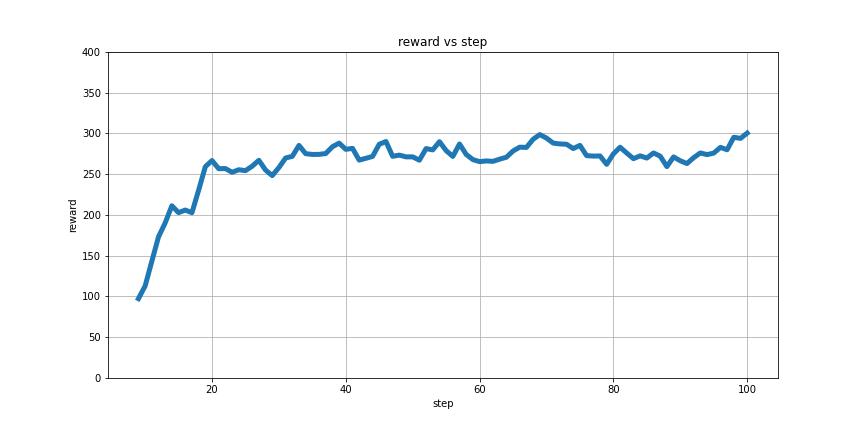}{
    The learning curve has reached a plateau and the performance has stablized, the training has converged.
}{
    YES
}
\analysisexemplar{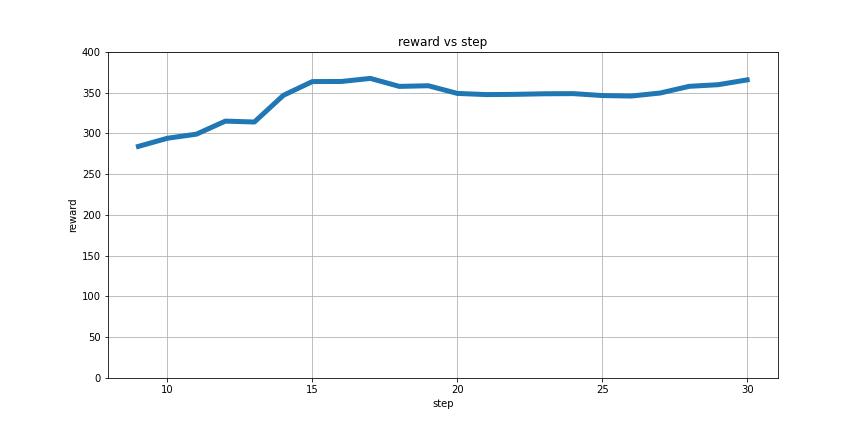}{
    The learning curve has reached a plateau and the performance has stablized, the training has converged.
}{
    YES
}
\analysisexemplar{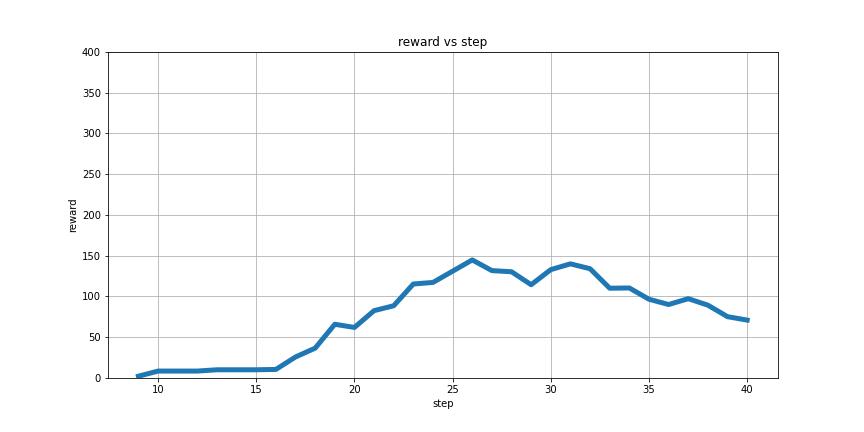}{
    The learning curve has has peaked an is degenerating again, the training had converged already.
}{
    YES
}
\analysisexemplar{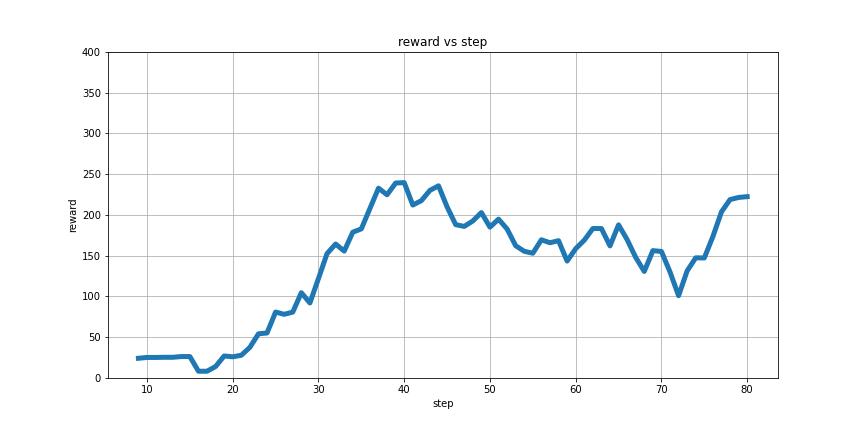}{
    The learning curve had degenerated after an earlier peak but is going up again, the training has not converged yet.
}{
    NO
}
\analysisexemplar{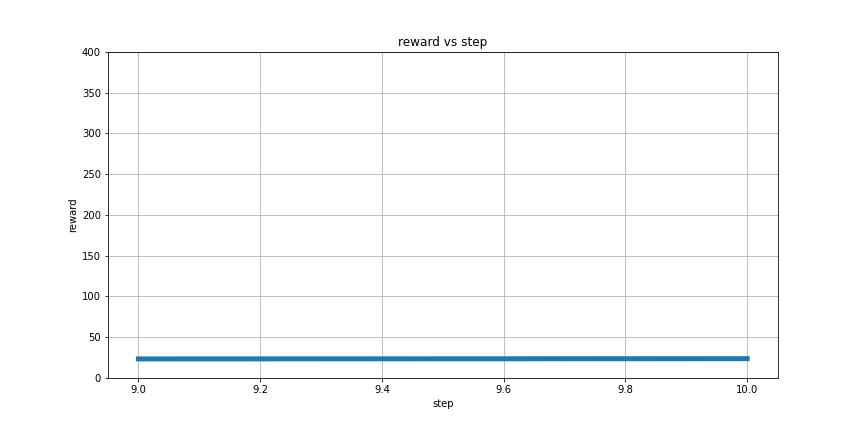}{
    There is no full learning curve yet, the training has not converged yet.
}{
    NO
}


\section{Response Examples}
\label{appendix:response_examples}

To illustrate the type of responses generated by the model, we present a number of examples from the beginning of a self-improvement run.

In the absence of successful trials, the very first proposal is to perform a tower building task.

\propositionexemplar{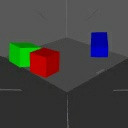}{
    [grasp anything, open gripper, close gripper, reach red, reach green, reach blue, above red, above green, above blue, lift red, lift green, lift blue, stack red on green, stack red on blue, stack green on red, stack green on blue, stack blue on red, stack blue on green]
}{
    []
}{
    I see three objects: red, green, blue. All possible spatial structures that can be built with them: a line where the three objects are placed next to each other; three dots where the three objects are apart from each other; a two-level tower with one object on top of another, and the third one on the side; a three-level tower with one object on top of another; a three-level slanted tower with one object on top of another but not aligned at the center. Previously, the robot has successfully built two-level towers, so trying to build a three-level tower should be both interesting and feasible to try now.
}{
    build a three-level tower with blue on top of green on top of red
}

Decomposition of this task illustrates the potential impact of temperature of the model; if we sample multiple decompositions in the same situation (but with different images) with a high temperature of $1.0$, the model produces plans of differing complexity.

\decompositionexemplar{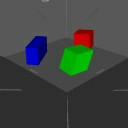}{
    build a three-level tower with blue on top of red on top of green
}{
    [grasp anything, open gripper, close gripper, reach red, reach green, reach blue, above red, above green, above blue, lift red, lift green, lift blue, stack red on green, stack red on blue, stack green on red, stack green on blue, stack blue on red, stack blue on green]
}{
    All three objects are on the bottom of the basket. To build a three-level tower with blue on top of red on top of green, I need to first stack red on green, then stack blue on red.
}{
    [reach green, above green, grasp anything, close gripper, lift green, reach red, above red, stack green on red, open gripper, reach blue, above blue, grasp anything, close gripper, lift blue, reach red, above red, stack blue on red, open gripper]
}

\decompositionexemplar{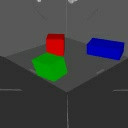}{
    build a three-level tower with blue on top of red on top of green
}{
    [grasp anything, open gripper, close gripper, reach red, reach green, reach blue, above red, above green, above blue, lift red, lift green, lift blue, stack red on green, stack red on blue, stack green on red, stack green on blue, stack blue on red, stack blue on green]
}{
    All three objects are on the bottom of the basket. To build a three-level tower with blue on top of red on top of green, I need to first stack red on green, then stack blue on red.
}{
    [stack red on green, stack blue on red]
}

The same task is also proposed later, after collecting several trials, with the reasoning that it failed several times and should thus be practiced. Note that if the same trial is attempted multiple times, it is combined and denoted with a success count (\texttt{x3}) for brevity; in the actual prompt, these are expanded into a list.

\propositionexemplar{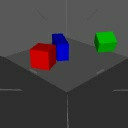}{
    [grasp anything, open gripper, close gripper, reach red, reach green, reach blue, above red, above green, above blue, lift red, lift green, lift blue, stack red on green, stack red on blue, stack green on red, stack green on blue, stack blue on red, stack blue on green, build a three-level tower with blue on top of green on top of red (x2), build a three-level tower with blue on top of red on top of green (x2)]
}{
    [build a three-level tower with blue on top of green on top of red (x2), build a three-level tower with blue on top of red on top of green (x5)]
}{
    I see three objects: red, green, blue. All possible spatial structures that can be built with them: a line where the three objects are placed next to each other; three dots where the three objects are apart from each other; a two-level tower with one object on top of another, and the third one on the side; a three-level tower with one object on top of another; a three-level slanted tower with one object on top of another but not aligned at the center. The robot has successfully built three-level towers but failed many times, so it needs more practice.
}{
    build a three-level tower with blue on top of red on top of green
}

A common failure case involves the model incorrectly injecting a description of the scene into the answer field of the response. This leads to a failure in the OneTwo library to parse the response even when it contains a task proposal, which in turn will cause the plan to be rejected.

\decompositionexemplar{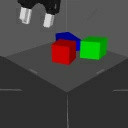}{
    build a two-level slanted tower with blue on top of red
}{
    [grasp anything, open gripper, close gripper, reach red, reach green, reach blue, above red, above green, above blue, lift red, lift green, lift blue, stack red on green, stack red on blue, stack green on red, stack green on blue, stack blue on red, stack blue on green]
}{
    All three objects are on the bottom of the basket. The robot can stack blue on red to build a two-level tower, but it might not be slanted. To make it slanted, the robot can first lift the red object up, then move above blue and drop it.
}{
    All three objects are on the bottom of the basket
}


\section{PAC Skills and Reward Functions}
\label{appendix:skill_rewards}

The reward functions used to train the PAC policy generally follow those used in \citet{bousmalis2023robocat}.
However, we use lower-level skills: instead of the strongly shaped and staged "stack and leave" reward they use, we only use its atomic components, as we would hope for our system to "discover" the heavily engineered, human-provided composite function in prior work. Thus our skill library consists of the following (where \texttt{X}, \texttt{Y} and \texttt{Z} are placeholders for all possible permutations of red, blue and green):
\begin{description}
    \item[\texttt{open\_gripper}:] Shaped; 0 if the gripper is closed, 1 if is maximally opened.
    \item[\texttt{close\_gripper}:] Shaped;  Inverse of \skill{open\_gripper}.
    \item[\texttt{grasp\_anything}:] Binary; 1 if the gripper's grasp sensor is triggered, 0 otherwise.
    \item[\texttt{reach\_X}:] Shaped; tangentially decaying distance between the robot's TCP and the center of object \texttt{X}.
    \item[\texttt{above\_X}:] Shaped; tangentially decaying distance between the robot's TCP and a point $10cm$ above the center of \texttt{X}.
    \item[\texttt{lift\_X}:] Shaped; 0 if the center of \texttt{X} is less than $5cm$ above the workspace surface, 1 if more than $10cm$ above, linearly interpolated between those limits.
    \item[\texttt{place\_X\_Y}:] Shaped; tangentially decaying distance between the center of  \texttt{X} and a point $4cm$ above \texttt{Y}.
    \item[\texttt{stack\_X\_Y}:] Shaped; \skill{place\_X\_Y}, but set to 0 if \skill{grasp\_anything} is non-zero.
\end{description}
During the self-improvement experiments, we add three more skills, composed of the above:
\begin{description}
    \item[\texttt{triple\_stack\_X\_Y\_Z}:] Product of \skill{stack\_X\_Y} and \skill{stack\_Y\_Z}.
    \item[\texttt{pyramid\_X\_Y\_Z}:] Product of \skill{stack\_X\_Y} and \skill{stack\_X\_Z}.
    \item[\texttt{inverse\_pyramid\_X\_Y\_Z}:] Product of \skill{stack\_X\_Y} and \skill{stack\_Z\_Y}.
\end{description}


\section{Progressive Skill Addition}
\label{appendix:progressive_skills}

\newcommand{\gradualproposition}[1]{
\begin{titlegraybox}{Proposition:}
\texttt{\textbf{Image observation of the current workspace:} }\includegraphics[width=0.15\textwidth]{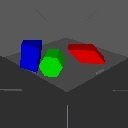}
\\\texttt{\textbf{Completed tasks so far:} #1}
\\\texttt{\textbf{Failed tasks that are too hard:} []} 
\end{titlegraybox}
}

\newcommand{\gradualpropositionresponse}[2]{
\begin{titlegraybox}{Proposition response:}
\texttt{\textbf{Reasoning:} #1} 
\\\texttt{\textbf{A:} #2}
\end{titlegraybox}
}

\newcommand{\gradualdecomposition}[2]{
\begin{titlegraybox}{Decomposition:}
\texttt{\textbf{Q:} #1
\\\textbf{Image observation of the current workspace:} }\includegraphics[width=0.15\textwidth]{exemplars/gradual.jpg}
\\\texttt{\textbf{Available skills:} #2}
\end{titlegraybox}

}

\newcommand{\gradualdecompositionresponse}[2]{
\begin{titlegraybox}{Decomposition response:}
\texttt{\textbf{Reasoning:} #1} 
\\\texttt{\textbf{A:} #2}
\end{titlegraybox}
}

\newcommand{\gradualretrieval}[2]{
\begin{titlegraybox}{Retrieval:}
\texttt{\textbf{Q:} #1}
\\\texttt{\textbf{Skill library:} #2}
\end{titlegraybox}
}

\newcommand{\gradualretrievalresponse}[2]{
\begin{titlegraybox}{Retrieval response:}
\texttt{\textbf{Reasoning:} #1} 
\\\texttt{\textbf{A:} #2}
\end{titlegraybox}
}

Below are the detailed model responses when examining proposal and decomposition at different steps of the pre-training process in section \ref{sec:progressive_skills}.
We note that since this is a post-hoc analysis, as the system had all skills available during the data collection experiments we performed.
Thus we can not collect \texttt{successful\_trials} and \texttt{failed\_trials} for proposition. Therefore for all the following rounds,
we use all skills available at the corresponding training step in the skill library as the \texttt{successful\_trials} and use an empty list as the \texttt{failed\_trials} for proposition.

\subsection{200k Learner Steps}

\begin{titlegraybox}{Skill library:}
\texttt{\\open gripper
\\close gripper
\\above red
\\above green
\\above blue
\\reach red
}
\end{titlegraybox}

\gradualproposition{
    [open gripper, close gripper, above red, above green, above blue, reach red]
}

\gradualpropositionresponse{
    I see three objects: red, green, blue. All possible spatial structures that can be built with them: a line where the three objects are placed next to each other; three dots where the three objects are apart from each other; a two-level tower with one object on top of another; a three-level tower with one object on top of another; a two-level slanted tower with one object on top of another but not aligned at the center; a three-level slanted tower with one object on top of another but not aligned at the center. Previously, the robot has successfully reached the red object but has not grasped any object yet, so trying to grasp an object should be both interesting and feasible to try now.
}{
    grasp the green object
}

\gradualdecomposition{
    grasp the green object
}{
    [open gripper, close gripper, above red, above green, above blue, reach red]
}

\gradualdecompositionresponse{
    The robot does not know how to grasp the green object directly. However, it can move its gripper above the green object and then close the gripper, which should result in grasping the green object.
}{
    [above green, close gripper]
}

\gradualretrieval{
    above green 
}{
    [open gripper, close gripper, above red, above green, above blue, reach red]
}

\gradualretrievalresponse{
    The query concerns with a configuration where something is above the green object. All skills except above green are irrelevant.
}{
    above green
}

\gradualretrieval{
    close gripper 
}{
    [open gripper, close gripper, above red, above green, above blue, reach red]
}

\gradualretrievalresponse{
    The query skill concerns with the gripper only. The second entry in the library is the only one that concerns with the gripper.
}{
    close gripper
}

\subsection{500k Learner Steps}

\begin{titlegraybox}{Skill library (added on top of 200k steps):}
\texttt{\\reach green
\\reach blue
\\grasp anything
\\lift red
}
\end{titlegraybox}

\gradualproposition{
    [open gripper, close gripper, above red, above green, above blue, reach red, reach green, reach blue, grasp anything, lift red]
}

\gradualpropositionresponse{
    I see three objects: red, green, blue. All possible spatial structures that can be built with them: a line where the three objects are placed next to each other; three dots where the three objects are apart from each other; a two-level tower with one object on top of another, and the third one on the side; a three-level tower with one object on top of another; a three-level slanted tower with one object on top of another but not aligned at the center. Previously, the robot has successfully manipulated the red object but not the green and blue ones, so trying to grasp the green object should be both interesting and feasible to try now.
}{
    grasp the green object
}

\gradualdecomposition{
    grasp the green object
}{
    [open gripper, close gripper, above red, above green, above blue, reach red, reach green, reach blue, grasp anything, lift red]
}

\gradualdecompositionresponse{
    The robot can reach above the green object, reach the green object, and grasp anything. Therefore, the robot can grasp the green object.
}{
    [The robot can reach above the green object, reach the green object, and grasp anything]
}

\gradualretrieval{
    The robot can reach above the green object
}{
    [open gripper, close gripper, above red, above green, above blue, reach red, reach green, reach blue, grasp anything, lift red]
}

\gradualretrievalresponse{
    The query concerns the robots capability of reaching a location above the green object. All skills in the library except above green are about robot actions, while above green describes a location relative to the green object.
}{
    above green
}

\gradualretrieval{
    reach the green object
}{
    [open gripper, close gripper, above red, above green, above blue, reach red, reach green, reach blue, grasp anything, lift red]
}

\gradualretrievalresponse{
    The query asks for reaching the green object. All skills except reach green are either irrelevant or describe a different object.
}{
    reach green
}

\gradualretrieval{
    and grasp anything
}{
    [open gripper, close gripper, above red, above green, above blue, reach red, reach green, reach blue, grasp anything, lift red]
}

\gradualretrievalresponse{
    The query skill does not induce any specific object configuration but asks for a grasping action. The only skill in the library that involves grasping is grasp anything.
}{
    grasp anything
}

\subsection{800k Learner Steps}

\begin{titlegraybox}{Skill library (added on top of 500k steps):}
\texttt{\\hold red over green
\\hold red over blue
\\hold green over blue
\\hold green over red
\\hold blue over red
\\hold blue over green
\\stack red on green
\\stack red on blue
\\stack green on blue
\\stack green on red
\\stack blue on red
\\stack blue on green
}
\end{titlegraybox}

\gradualproposition{
    [open gripper, close gripper, above red, above green, above blue, reach red, reach green, reach blue, grasp anything, lift red, hold red over green, hold red over blue, hold green over blue, hold green over red, hold blue over red, hold blue over green, stack red on green, stack red on blue, stack green on blue, stack green on red, stack blue on red, stack blue on green]
}

\gradualpropositionresponse{
    I see three objects: red, green, blue. All possible spatial structures that can be built with them: a line where the three objects are placed next to each other; three dots where the three objects are apart from each other; a two-level tower with one object on top of another, and the third object next to the tower; a three-level tower with one object on top of another; a three-level slanted tower with one object on top of another but not aligned at the center. The robot has successfully built all two-level towers, so trying to build a three-level tower should be both interesting and feasible to try now.
}{
    build a three-level tower with blue on top of red on top of green
}

\gradualdecomposition{
    build a three-level tower with blue on top of red on top of green
}{
    [open gripper, close gripper, above red, above green, above blue, reach red, reach green, reach blue, grasp anything, lift red, hold red over green, hold red over blue, hold green over blue, hold green over red, hold blue over red, hold blue over green, stack red on green, stack red on blue, stack green on blue, stack green on red, stack blue on red, stack blue on green]
}

\gradualdecompositionresponse{
    All three objects are on the bottom of the basket. To build a three-level tower with blue on top of red on top of green, I need to stack red on green first, then stack blue on red.
}{
    [stack red on green, stack blue on red]
}

\gradualretrieval{
    stack red on green
}{
    [open gripper, close gripper, above red, above green, above blue, reach red, reach green, reach blue, grasp anything, lift red, hold red over green, hold red over blue, hold green over blue, hold green over red, hold blue over red, hold blue over green, stack red on green, stack red on blue, stack green on blue, stack green on red, stack blue on red, stack blue on green]
}

\gradualretrievalresponse{
    The query asks for a specific object configuration where the red object is on top of the green one. All skills except those related to stack only involve individual object manipulation. The desired configuration is specifically achieved by stack red on green.
}{
    stack red on green
}

\gradualretrieval{
    stack blue on red
}{
    [open gripper, close gripper, above red, above green, above blue, reach red, reach green, reach blue, grasp anything, lift red, hold red over green, hold red over blue, hold green over blue, hold green over red, hold blue over red, hold blue over green, stack red on green, stack red on blue, stack green on blue, stack green on red, stack blue on red, stack blue on green]
}

\gradualretrievalresponse{
    The query asks for a specific object configuration - blue on top of red. All other skills either manipulate a single object or create different configurations. Only stack blue on red matches the desired outcome.
}{
    stack blue on red
}

\subsection{After Self-improvement}

\begin{titlegraybox}{Skill library (added on top of 800k steps):}
\texttt{stack green on blue and red on green
\\stack blue on green and red on blue
\\stack red on green and blue on red
\\stack green on red and blue on green
\\stack blue on red and green on blue
\\stack red on blue and green on red
\\build a pyramid with red on top and green and blue at the bottom
\\build a pyramid with red on top and blue and green at the bottom
\\build a pyramid with green on top and blue and red at the bottom
\\build a pyramid with green on top and red and blue at the bottom
\\build a pyramid with blue on top and red and green at the bottom
\\build a pyramid with blue on top and green and red at the bottom
\\build an inverted pyramid with green and blue at the top and red at the bottom
\\build an inverted pyramid with blue and green at the top and red at the bottom
\\build an inverted pyramid with blue and red at the top and green at the bottom
\\build an inverted pyramid with red and blue at the top and green at the bottom
\\build an inverted pyramid with red and green at the top and blue at the bottom
\\build an inverted pyramid with green and red at the top and blue at the bottom
}
\end{titlegraybox}

\gradualproposition{
    [open gripper, close gripper, above red, above green, above blue, reach red, reach green, reach blue, grasp anything, lift red, hold red over green, hold red over blue, hold green over blue, hold green over red, hold blue over red, hold blue over green, stack red on green, stack red on blue, stack green on blue, stack green on red, stack blue on red, stack blue on green, stack green on blue and red on green, stack blue on green and red on blue, stack red on green and blue on red, stack green on red and blue on green, stack blue on red and green on blue, stack red on blue and green on red, build a pyramid with red on top and green and blue at the bottom, build a pyramid with red on top and blue and green at the bottom, build a pyramid with green on top and blue and red at the bottom, build a pyramid with green on top and red and blue at the bottom, build a pyramid with blue on top and red and green at the bottom, build a pyramid with blue on top and green and red at the bottom, build an inverted pyramid with green and blue at the top and red at the bottom, build an inverted pyramid with blue and green at the top and red at the bottom, build an inverted pyramid with blue and red at the top and green at the bottom, build an inverted pyramid with red and blue at the top and green at the bottom, build an inverted pyramid with red and green at the top and blue at the bottom, build an inverted pyramid with green and red at the top and blue at the bottom]
}

\gradualpropositionresponse{
    I see three objects: red, green, blue. All possible spatial structures that can be built with them: a line where the three objects are placed next to each other; three dots where the three objects are apart from each other; a two-level tower with two objects on top of one another; a three-level tower with three objects on top of one another; a two-level slanted tower with two objects on top of one another but not aligned at the center; a three-level slanted tower with three objects on top of one another but not aligned at the center. The robot has already built all possible two-level and three-level towers, including slanted ones. The robot has not yet built a line with all three objects.
}{
    build a line with the red object next to the green object and the blue object next to the green object
}

\gradualdecomposition{
    build a line with the red object next to the green object and the blue object next to the green object
}{
    [open gripper, close gripper, above red, above green, above blue, reach red, reach green, reach blue, grasp anything, lift red, hold red over green, hold red over blue, hold green over blue, hold green over red, hold blue over red, hold blue over green, stack red on green, stack red on blue, stack green on blue, stack green on red, stack blue on red, stack blue on green, stack green on blue and red on green, stack blue on green and red on blue, stack red on green and blue on red, stack green on red and blue on green, stack blue on red and green on blue, stack red on blue and green on red, build a pyramid with red on top and green and blue at the bottom, build a pyramid with red on top and blue and green at the bottom, build a pyramid with green on top and blue and red at the bottom, build a pyramid with green on top and red and blue at the bottom, build a pyramid with blue on top and red and green at the bottom, build a pyramid with blue on top and green and red at the bottom, build an inverted pyramid with green and blue at the top and red at the bottom, build an inverted pyramid with blue and green at the top and red at the bottom, build an inverted pyramid with blue and red at the top and green at the bottom, build an inverted pyramid with red and blue at the top and green at the bottom, build an inverted pyramid with red and green at the top and blue at the bottom, build an inverted pyramid with green and red at the top and blue at the bottom]
}

\gradualdecompositionresponse{
    The robot can stack green on blue and red on green in one go, which will result in the desired configuration.
}{
    [stack green on blue and red on green]
}

\gradualretrieval{
    stack green on blue and red on green
}{
    [open gripper, close gripper, above red, above green, above blue, reach red, reach green, reach blue, grasp anything, lift red, hold red over green, hold red over blue, hold green over blue, hold green over red, hold blue over red, hold blue over green, stack red on green, stack red on blue, stack green on blue, stack green on red, stack blue on red, stack blue on green, stack green on blue and red on green, stack blue on green and red on blue, stack red on green and blue on red, stack green on red and blue on green, stack blue on red and green on blue, stack red on blue and green on red, build a pyramid with red on top and green and blue at the bottom, build a pyramid with red on top and blue and green at the bottom, build a pyramid with green on top and blue and red at the bottom, build a pyramid with green on top and red and blue at the bottom, build a pyramid with blue on top and red and green at the bottom, build a pyramid with blue on top and green and red at the bottom, build an inverted pyramid with green and blue at the top and red at the bottom, build an inverted pyramid with blue and green at the top and red at the bottom, build an inverted pyramid with blue and red at the top and green at the bottom, build an inverted pyramid with red and blue at the top and green at the bottom, build an inverted pyramid with red and green at the top and blue at the bottom, build an inverted pyramid with green and red at the top and blue at the bottom]
}

\gradualretrievalresponse{
    The query asks for a three-object configuration with green on blue and red on green. This configuration is exactly what the 22nd skill in the library describes.
}{
    stack green on blue and red on green
}

\end{document}